\documentclass[10pt,twocolumn,letterpaper]{article}

\usepackage[pagenumbers]{cvpr} 

\usepackage{color, colortbl}
\usepackage{xcolor}
\usepackage{xspace}
\usepackage{multirow}
\usepackage{adjustbox}
\usepackage[accsupp]{axessibility}  
\usepackage{color, colortbl}
\usepackage{xcolor}
\usepackage{xspace}
\usepackage{multirow}
\definecolor{Gray}{gray}{0.9}

\usepackage{makecell}
\usepackage{algorithmic}
\usepackage{algorithm}
\usepackage{afterpage}

\definecolor{Gray}{gray}{0.9}

\definecolor{cvprblue}{rgb}{0.21,0.49,0.74}
\usepackage[pagebackref,breaklinks,colorlinks,allcolors=cvprblue]{hyperref}

\def\paperID{12865} 
\def\confName{CVPR}
\def\confYear{2025}

\title{Subnet-Aware Dynamic Supernet Training for Neural Architecture Search}

\author{Jeimin Jeon$^{1,2}$ \quad\quad\quad Youngmin Oh$^{1}$ \quad\quad\quad Junghyup Lee$^{3}$ \quad\quad\quad Donghyeon Baek$^{1}$ \vspace*{0.2mm}\\
Dohyung Kim$^{4}$ \quad\quad\quad Chanho Eom$^{5}$ \quad\quad\quad Bumsub Ham$^{1}$\thanks{Corresponding author.} \vspace*{3mm}\\
$^{1}$Yonsei University \quad\quad\quad $^{2}$Articron Inc. \quad\quad\quad $^{3}$Samsung Research \\
$^{4}$Samsung Advanced Institute of Technology \quad\quad\quad $^{5}$Chung-Ang University \\
{ \url{https://cvlab.yonsei.ac.kr/projects/DYNAS/}}
}

\begin{document}
\maketitle
\begin{abstract}
    \textit{N}-shot neural architecture search~(NAS) exploits a supernet containing all candidate subnets for a given search space. The subnets are typically trained with a static training strategy~(\eg, using the same learning rate~(LR) scheduler and optimizer for all subnets). This however does not consider that individual subnets have distinct characteristics, leading to two problems: (1)~The supernet training is biased towards the low-complexity subnets~(unfairness); (2)~the momentum update in the supernet is noisy~(noisy momentum). We present a dynamic supernet training technique to address these problems by adjusting the training strategy adaptive to the subnets. Specifically, we introduce a complexity-aware LR scheduler~(CaLR) that controls the decay ratio of LR adaptive to the complexities of subnets, which alleviates the unfairness problem. We also present a momentum separation technique~(MS). It groups the subnets with similar structural characteristics and uses a separate momentum for each group, avoiding the noisy momentum problem. Our approach can be applicable to various \textit{N}-shot NAS methods with marginal cost, while improving the search performance drastically. We validate the effectiveness of our approach on various search spaces (\eg, NAS-Bench-201, Mobilenet spaces) and datasets (\eg, CIFAR-10/100, ImageNet).
  \end{abstract} 

  \begin{figure}[t]
    \captionsetup{font={small}}
    \begin{center}
       \begin{subfigure}{\columnwidth}
          \centering
          \captionsetup{justification=centering}
          \subcaptionbox{Unfairness.}{
          \includegraphics[width=0.48\textwidth]{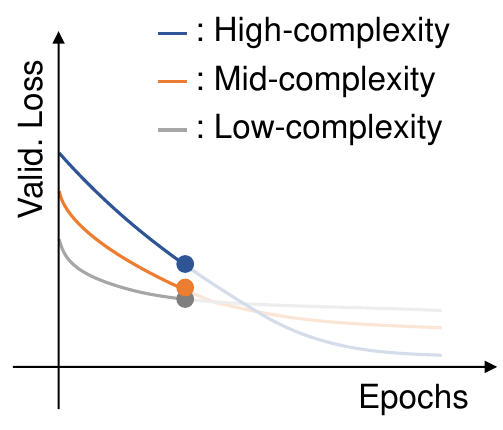}}
          \subcaptionbox{Noisy Momentum.}{
           \includegraphics[width=0.48\textwidth]{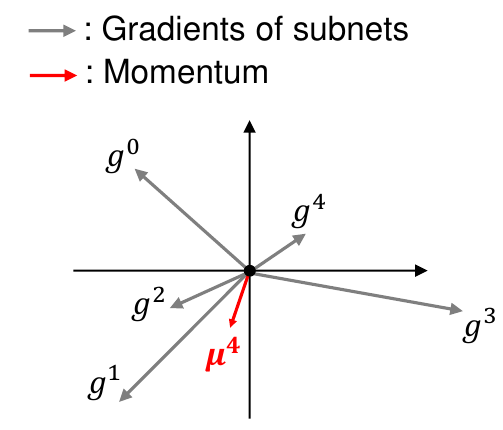}}
       \end{subfigure}
    \end{center}
    \vspace{-0.6cm}
    \caption{Illustrations of the challenges of \textit{N}-shot NAS methods. (a)~We visualize validation losses for the subnets having different complexities at training time. Existing methods do not consider the distinct optimization speed of subnets \wrt complexities. This causes an unfairness problem, where the high-complexity subnet is trained insufficiently, and the predicted performance falls behind the low-complexity one, even if it might be supposed to provide better performance. (b)~We illustrate gradients $g^t$ of subnets and the momentum $\mu^t$ at $t$-th iteration. We can see that the gradients vary according to the subnets, resulting in a noisy momentum and preventing a stable training process. (Best viewed in color.)}
    \label{fig:teaser}
    \vspace{-0.6cm}
  \end{figure}

\section{Introduction}
\label{sec:intro}
Designing network architectures customized for each hardware configuration has led to a significant attention over the last decade, with the increasing demand for applying them to various platforms. Neural architecture search~(NAS) enables finding optimal network architectures automatically, providing better performance, compared to hand-crafted ones. Early works on NAS exploit reinforcement learning~\cite{williams1992simple} or evolutionary algorithms~\cite{goldberg1991comparative} for searching network architectures, typically requiring extensive computational resources,~\eg,~in the order of thousands of GPU days~\cite{zoph2016neural,real2019regularized}, which limits the practical applicability.

To reduce the search cost, \textit{N}-shot NAS methods~\cite{guo2020single,peng2021pi,chu2021fairnas,ha2021sumnas,zhao2021few,hu2022generalizing, liu2018darts, chen2019progressive,chu2020darts, ye2022b,chu2020fairdarts} have recently been introduced. They exploit either a single supernet for one-shot NAS~\cite{guo2020single,peng2021pi,chu2021fairnas,ha2021sumnas} or multiple sub-supernets partitioned from a supernet for few-shot NAS~\cite{zhao2021few,hu2022generalizing}. The supernet is an over-parameterized network that contains all candidate architectures~(\ie, subnets) in a search space, serving as a performance predictor of subnets. For training the supernets, sampling-based methods~\cite{guo2020single,peng2021pi,chu2021fairnas,ha2021sumnas} are widely adopted nowadays due to their efficiency. These methods optimize the supernet by training one or multiple subnets sampled at each iteration. Existing methods, however, train the sampled subnets with a static training strategy, employing a fixed learning rate~(LR) scheduler and the same optimizer for all subnets within the supernet. These methods overlook the fact that the subnets have different characteristics (\eg,~complexities and structures) from each other, and thus they should be trained in different settings. The static training strategy mainly has two problems. First, it causes an unfairness problem, where the supernet training is biased towards the low-complexity subnets. The subnets with higher complexity typically need more training iterations than low-complexity counterparts, to reach sufficient performance levels for accurate ranking. Discarding this difference in the supernet training process places the rank of the high-complexity subnets behind the low-complexity ones, even when they might be supposed to provide better performance~(Fig.~\ref{fig:teaser}(a)). Second, sharing a single optimizer for all subnets causes a noisy momentum problem. That is, the momentum storing the running mean of gradients becomes noisy, due to varying gradients across the subnets. Generally, exploiting a momentum in an optimizer stabilizes the training process, as it mitigates oscillations of network parameters by using the historical gradients. However, in the case of the supernet, the gradients come from diverse subnets within the search space, and they can vary significantly~(Fig.~\ref{fig:teaser}(b)). Accordingly, the momentum characterized by the accumulation of the gradients fails to represent the gradients of individual subnets accurately, which might cause the gradients to steer toward the wrong direction.

In this paper, we introduce a novel dynamic supernet training framework that exploits distinct training strategies adaptive to individual subnets, effectively addressing the aforementioned problems for training supernets. To this end, we propose a complexity-aware LR scheduler~(CaLR) that alleviates the unfairness problem for the subnets. Specifically, considering the number of parameters to tune for each subnet, we slowly decay LR of high-complexity subnets to encourage more parameter updates, while we accelerate the LR decay for low-complexity ones. This enables a more balanced training of each subnet according to its complexity, alleviating the ranking inconsistencies. Moreover, we introduce two metrics, Complexity Bias~(CB) and Complexity-Convergence Correlation~($C^3$), that measure the unfairness problem in supernet training, and show that our CaLR can effectively reduce the unfairness problem. We also present a momentum separation technique~(MS) that stabilizes the supernet training process. Motivated by the observation that the subnets with similar structures provide similar gradients~\cite{kornblith2019similarity, peng2021pi}, we cluster the subnets based on the structural characteristics, and use a separate momentum buffer for each cluster, stabilizing the momentum updates. We show that our dynamic training strategy~(CaLR + MS) can be applied to various \textit{N}-shot NAS methods~\cite{guo2020single,chu2021fairnas,zhao2021few} efficiently, improving the search performance, and demonstrate the effectiveness of our method on various datasets, including CIFAR-10/100~\cite{krizhevsky2009cifar} and ImageNet~\cite{deng2009imagenet}. The main contributions of our work are summarized as follows:
\begin{itemize}[leftmargin=*]

  \item We propose CaLR adjusting the LR based on the complexity of the subnets, addressing the unfairness problem. To measure the unfairness problem in supernet training, we introduce evaluation metrics, CB and $C^3$.
  \item We propose MS exploiting a distinct momentum buffer for the subnets having similar structural characteristics, mitigating the noisy momentum problem and stabilizing a training process.
  \item Extensive experiments demonstrate that our approach enhances various \textit{N}-shot NAS methods consistently, with a negligible additional search time. 
\end{itemize}

\section{Related Work}

\begin{figure*}[t]
  \captionsetup{font={small}}
 \centering
 \begin{subfigure}{0.238\textwidth}
     \centering
     \includegraphics[width=\columnwidth]{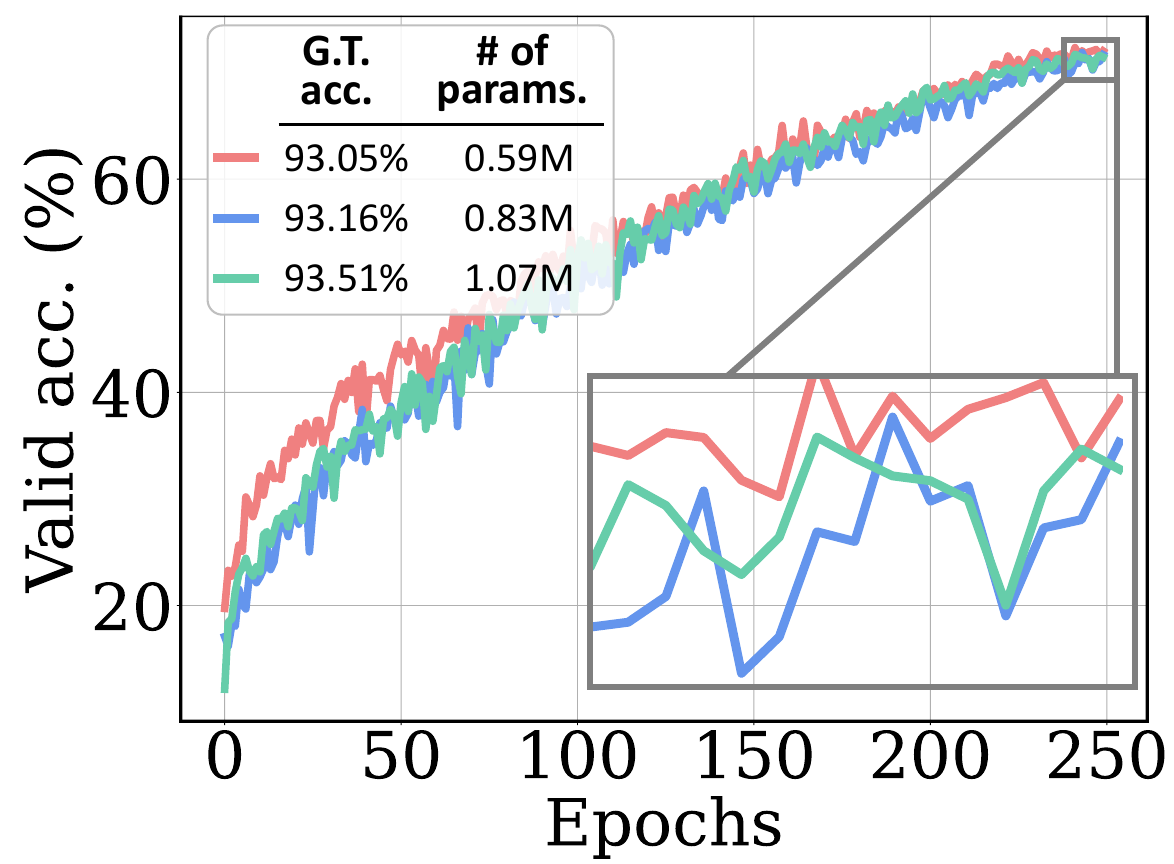}
     \caption{SPOS~\cite{guo2020single}.}
     \label{fig:spos}
 \end{subfigure}
 \hfill
 \begin{subfigure}{0.238\textwidth}
     \centering
     \includegraphics[width=\columnwidth]{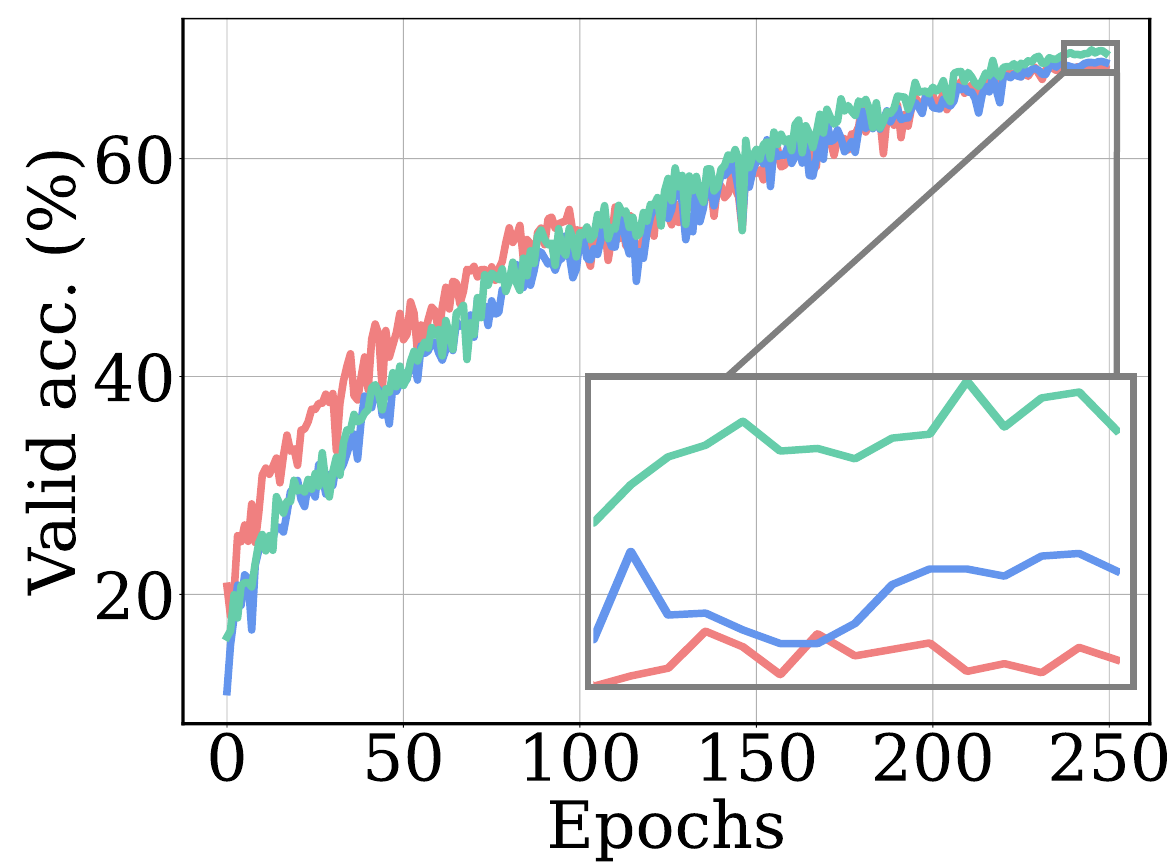}
     \caption{SPOS w/ CaLR.}
     \label{fig:spos_calr}
 \end{subfigure}
 \hfill
 \begin{subfigure}{0.227\textwidth}
     \centering
     \includegraphics[width=\columnwidth]{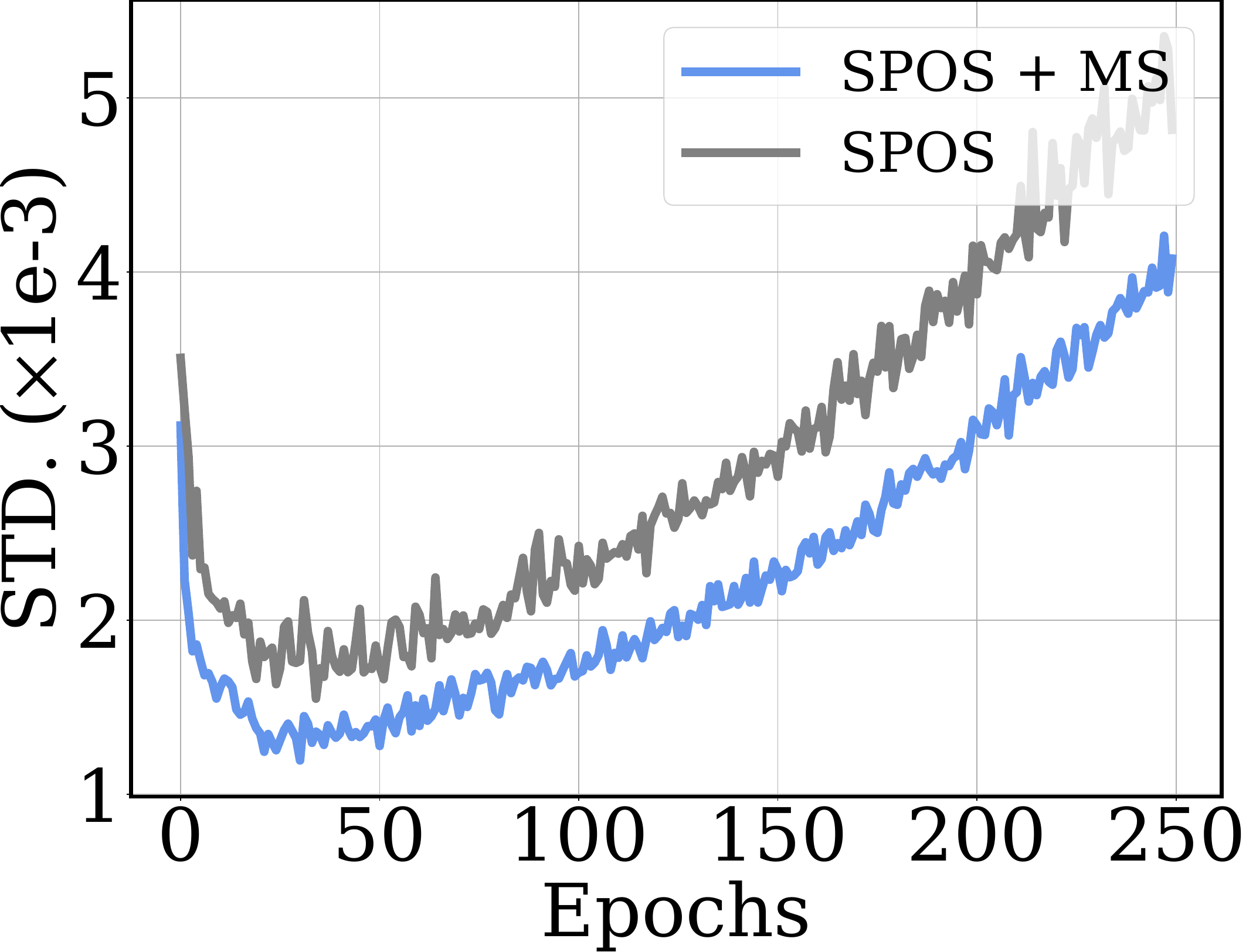}
     \caption{Gradient Consistency.}
     \label{fig:gradient_consistency}
 \end{subfigure}
 \hfill
 \begin{subfigure}{0.241\textwidth}
     \centering
     \includegraphics[width=\columnwidth]{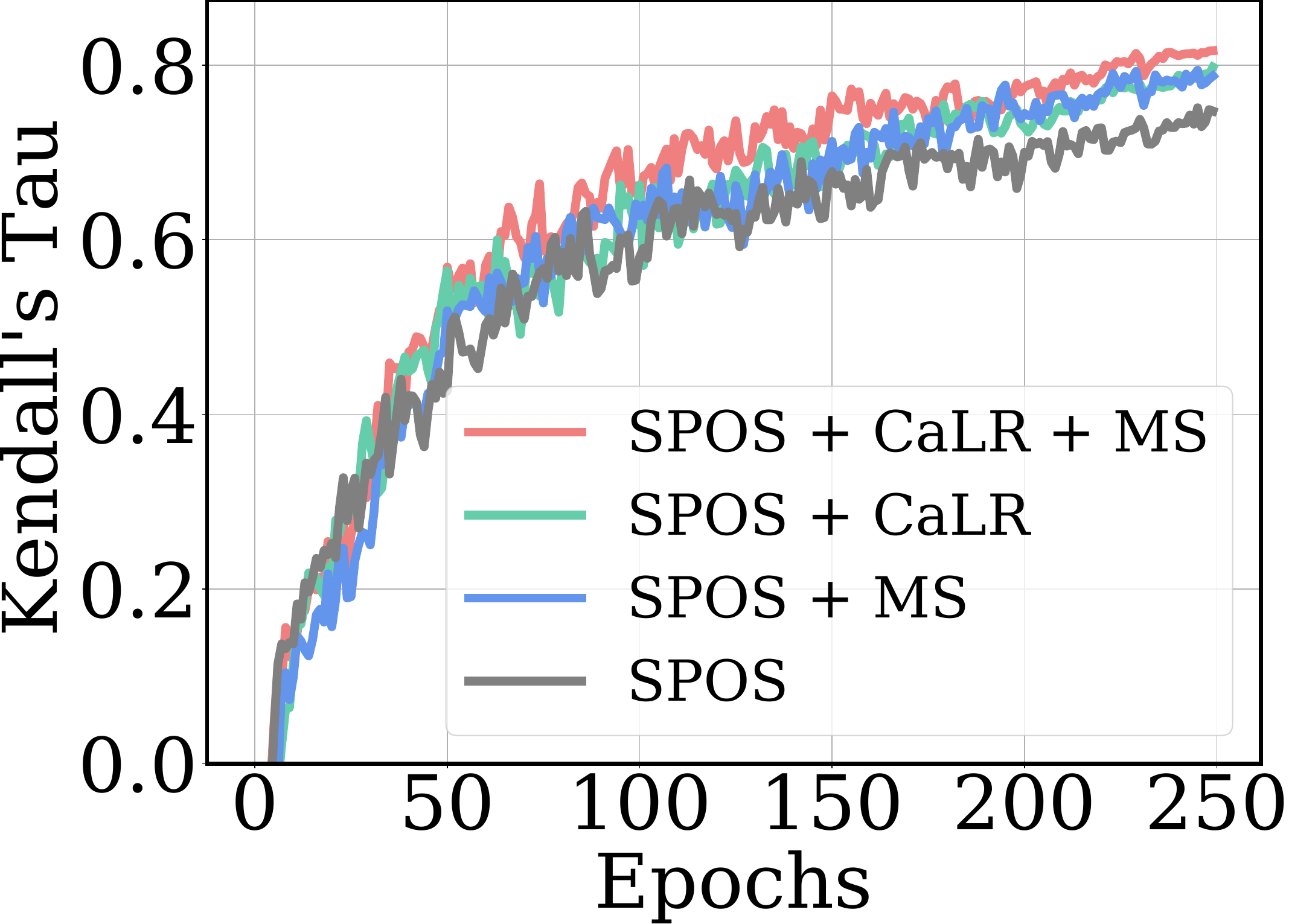}
     \caption{Ranking Consistency.}
     \label{fig:ranking_consistency}
 \end{subfigure}
 \vspace{-0.2cm}
 \caption{Empirical comparisons of SPOS~\cite{guo2020single} and SPOS with our dynamic training strategy. We train a supernet using the NAS-Bench-201 search space~\cite{dong2020bench} on CIFAR-10~\cite{krizhevsky2009cifar}. (a-b) Validation accuracies for three subnets sampled from the supernet using SPOS~\cite{guo2020single} without and with CaLR. Note that the sampled subnets have different complexities and ground-truth accuracies. (c) Plots of gradient consistency in terms of the standard deviation of gradients~\cite{lu2023pa}. A smaller value indicates more consistent gradient direction over the training iterations. (d) Comparisons of various methods in terms of ranking consistency of the supernet, using the Kendall's Tau~\cite{kendall1948rank}.}
 \label{fig:observation}
 \vspace{-0.5cm}
\end{figure*}

\paragraph{One-shot NAS.}
\label{subsec:one-shot}
Early approaches to NAS adopt reinforcement learning~\cite{williams1992simple} or evolutionary algorithms~\cite{goldberg1991comparative} to find an optimal network architecture, taking a huge amount of search costs~\cite{zoph2016neural,real2019regularized}. To tackle this problem, one-shot NAS methods~\cite{liu2018darts, chen2019progressive,chu2020darts, ye2022b,chu2020fairdarts, guo2020single,hu2020angle,chen2021one,peng2020cream,wang2021alphanet, chu2021fairnas,ha2021sumnas} exploit a supernet, that is, an over-parametrized network including all candidate subnets in a search space. Once training a single supernet, one-shot NAS methods can predict the performance of several subnets in the supernet without training each from scratch. Differentiable methods~\cite{liu2018darts, chen2019progressive,chu2020darts, ye2022b,chu2020fairdarts} formulate each layer as a weighted combination of candidate operations, where the weights are trained alternately with the parameters for the supernet. They should store gradients and activations for each operation at every layer, requiring substantial memory. To address this problem, sampling-based approaches~\cite{guo2020single,hu2020angle,chen2021one,peng2020cream,wang2021alphanet, chu2021fairnas,ha2021sumnas} optimize the supernet by training a randomly sampled subnet at each step, storing the gradients and activations for the single subnet only. Despite the efficiency, they suffer from a poor ranking consistency mainly due to the weight sharing among the subnets~\cite{yu2019evaluating}. That is, the weights of the subnets interfere with one another, which leads each subnet to converge insufficiently, and degrades the reliability of the supernet in terms of performance predictions. This problem can be alleviated by narrowing a search space~\cite{hu2020angle,chen2021one}, or exploiting distillation~\cite{peng2020cream,wang2021alphanet} or multi-subnet sampling strategies~\cite{chu2021fairnas, ha2021sumnas}. Specifically, the works of~\cite{hu2020angle,chen2021one} shrink the search space by removing the redundant operations in order to lower the degree of the sharing. In~\cite{peng2020cream,wang2021alphanet}, a distillation technique~\cite{hinton2015distilling} is adopted to boost the convergence of individual subnets. FairNAS~\cite{chu2021fairnas} and SUMNAS~\cite{ha2021sumnas} exploit multi-subnet sampling strategies for more balanced training across each subnet. Specifically, FairNAS~\cite{chu2021fairnas} samples multiple subnets for each training step, and trains them under the same setting. SUMNAS~\cite{ha2021sumnas} further improves FairNAS by adopting a meta-learning approach to alleviate the forgetting problem occurring during a supernet training~\cite{zhang2020overcoming}. Namely, the knowledge from subnets becomes forgotten gradually, as the training process goes on. These methods still exploit a static strategy to train a supernet, while not considering the unique characteristics of each subnet.

Recent approaches~\cite{you2020greedynas,huang2022greedynasv2,lu2023pa,zhang2023shiftnas} propose to use a non-uniform subnet sampling strategy, with an observation that sampling each subnet uniformly across the supernet with an equal probability is sub-optimal. GreedyNAS~\cite{you2020greedynas,huang2022greedynasv2} mainly samples potentially high-performing subnets, instead of weak ones. PA\&DA~\cite{lu2023pa} shows that reducing gradient variances in the supernet enhances the search performance in NAS, and proposes to sample subnets and images that lead to lower gradient variances. ShiftNAS~\cite{zhang2023shiftnas} uses more training steps for the subnets that are not sufficiently trained compared to others. Our method is orthogonal to these approaches in that we focus on the optimization process of each subnet, rather than the sampling strategies. Ours is also efficient to implement, while estimating the distributions for sampling is computationally expensive. 

\vspace{-0.3cm}
\paragraph{Few-shot NAS.}
\label{subsec:few-shot}
Few-shot NAS methods~\cite{zhao2021few,hu2022generalizing} divide a supernet into multiple sub-supernets, and apply one-shot NAS approaches for each sub-supernet, to prevent coupling weights among subnets, providing better results, compared to one-shot approaches. Few-shot NAS methods focus on designing criteria for partitioning the supernet, and various strategies have been introduced. FSNAS~\cite{zhao2021few} shows that randomly partitioning a supernet could be a simple solution to obtain decoupled weights among subnets. GMNAS~\cite{hu2022generalizing} improves the splitting strategy using the gradient similarity between the subnets. Specifically, it assumes that subnets with different gradient directions are more likely to interfere with each other, and thus they should be separated. Similar to the one-shot NAS approaches, our method can also be applicable to the few-shot NAS methods, helping to better optimize each sub-supernet.

\section{Method}
\label{sec:methods}
\vspace{-0.2cm}

In this section, we describe our approach within a framework of single-path one-shot NAS~(SPOS)~\cite{guo2020single}. Extensions to other approaches~\cite{chu2021fairnas,zhao2021few} other than SPOS are shown in the supplement. We first review the search process of SPOS~\cite{guo2020single}~(Sec.~\ref{subsec:preliminaries}), and describe our dynamic supernet training method~(Sec.~\ref{subsec:dstf}).
 
\subsection{Supernet training and architecture search}
\label{subsec:preliminaries}
One-shot NAS methods aim to train weights $\mathcal{W}$ of a supernet that contains all subnets $\alpha$ in a search space $\mathcal{A}$. They try to minimize an expected loss across all the subnets to estimate optimal weights~$\mathcal{W}^\star$ as follows: 
\begin{equation}
  \mathcal{W}^\star = \underset{\mathcal{W}}{\mathrm{argmin}} \ \mathbb{E}_{\substack{\alpha \sim \mathcal{A}}}[\mathcal{L}(\alpha;\mathcal{W}(\alpha))],
  \label{eq:supernet_train_obj}
\end{equation}
where we denote by $\mathcal{W}(\alpha)$ active weights for the subnet~$\alpha$. $\mathcal{L}(\alpha;\mathcal{W}(\alpha))$ is a training loss for the subnet~$\alpha$ with the weights of~$\mathcal{W}(\alpha)$. At each training step, a subnet $\alpha$ is sampled from the search space $\mathcal{A}$ uniformly, and the weights $\mathcal{W}(\alpha)$ are updated to minimize the training loss $\mathcal{L}(\alpha;\mathcal{W}(\alpha))$. Exploiting the trained supernet as a performance predictor for all possible subnets, we can search for the optimal architecture $\mathcal{\alpha}^\star$ under specific constraints, including FLOPs and latency, as follows:
\begin{equation} 
  \mathcal{\alpha}^\star = \underset{\alpha \in \mathcal{A}}{\mathrm{argmin}} \ {L}_{val}(\alpha;\mathcal{W}(\alpha)),
\end{equation}
where ${L}_{val}$ is a validation loss.



\subsection{Subnet-aware dynamic training}
\label{subsec:dstf}

Since the supernet is used to predict the performance of each subnet, improving a ranking consistency for the supernet is essential to achieve the better search performance. Existing methods assume that all subnets have the same convergence rate, if they are trained with the same setting. Accordingly, they typically use a static supernet training strategy, where the weights associated with each subnet~(\eg,~${W}(\alpha)$ and $\alpha$, respectively, in Eq.~\eqref{eq:supernet_train_obj}) are updated with the same LR scheduler and optimizer. The assumption does not hold in practice, since characteristics of the subnets are different from each other, leading to the unfairness and noisy momentum problems. To better understand these problems, we provide empirical analyses of the effect of the static supernet training strategy in Fig.~\ref{fig:observation}. First, we visualize the validation accuracies of subnets with different complexities at training time~(Fig.~\ref{fig:observation}(a)). We can see that a static training approach may underestimate the performance of high-complexity subnets (green line) compared to low-complexity ones (red line), even though the former potentially has higher ground-truth accuracy. We also plot the gradient variance of the supernet at each training epoch~(Fig.~\ref{fig:observation}(c)). We can see that the variance is large over the training process, as the gradients vary significantly. This can cause a noisy momentum problem, making a supernet training process unstable. These problems lead to lower ranking consistency~(Fig.~\ref{fig:observation}(d)), which can result in sub-optimal search performance.

To address these problems, we introduce a novel supernet training method that adaptively adjusts training strategies for each subnet. Specifically, we propose a complexity-aware LR scheduler~(CaLR), the LR scheduler adjusting LR values adaptively \wrt complexities of the subnets, to alleviate the unfairness problem. We also present a momentum separation technique~(MS) that assigns different momentum buffers to the subnets, while considering their structural characteristics, stabilizing a training process. In the following, we provide detailed descriptions of CaLR and MS. 

\begin{figure}
    \captionsetup{font={small}}
    \centering
    \begin{subfigure}{0.47\linewidth}
        \centering
        \includegraphics[width=0.99\columnwidth]{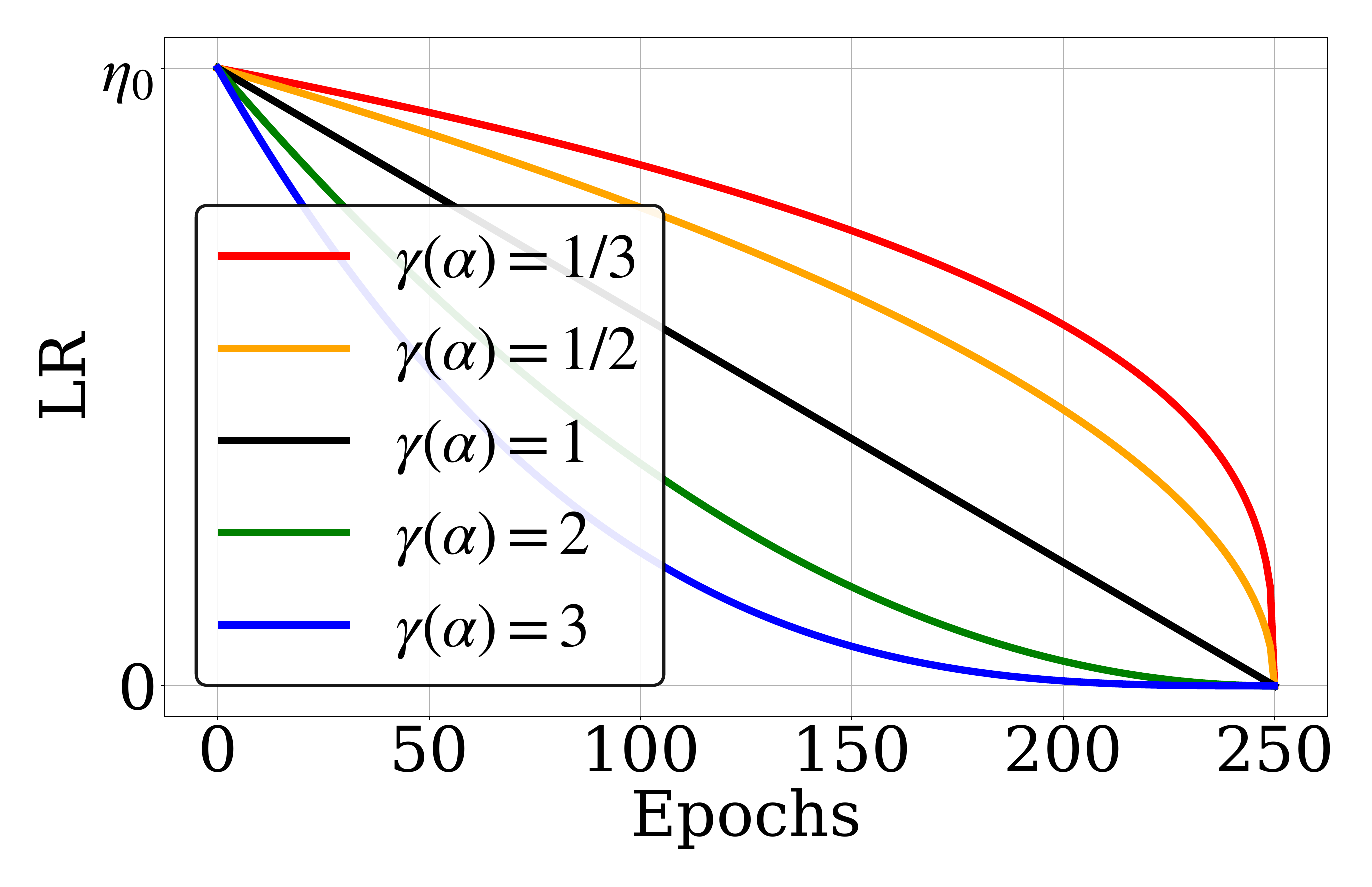}
        \vskip -0.1in
        \caption{}
        \label{fig:adaptive_lr}
    \end{subfigure}
    \hfill
    \begin{subfigure}{0.49\linewidth}
        \centering
        \includegraphics[width=0.99\columnwidth]{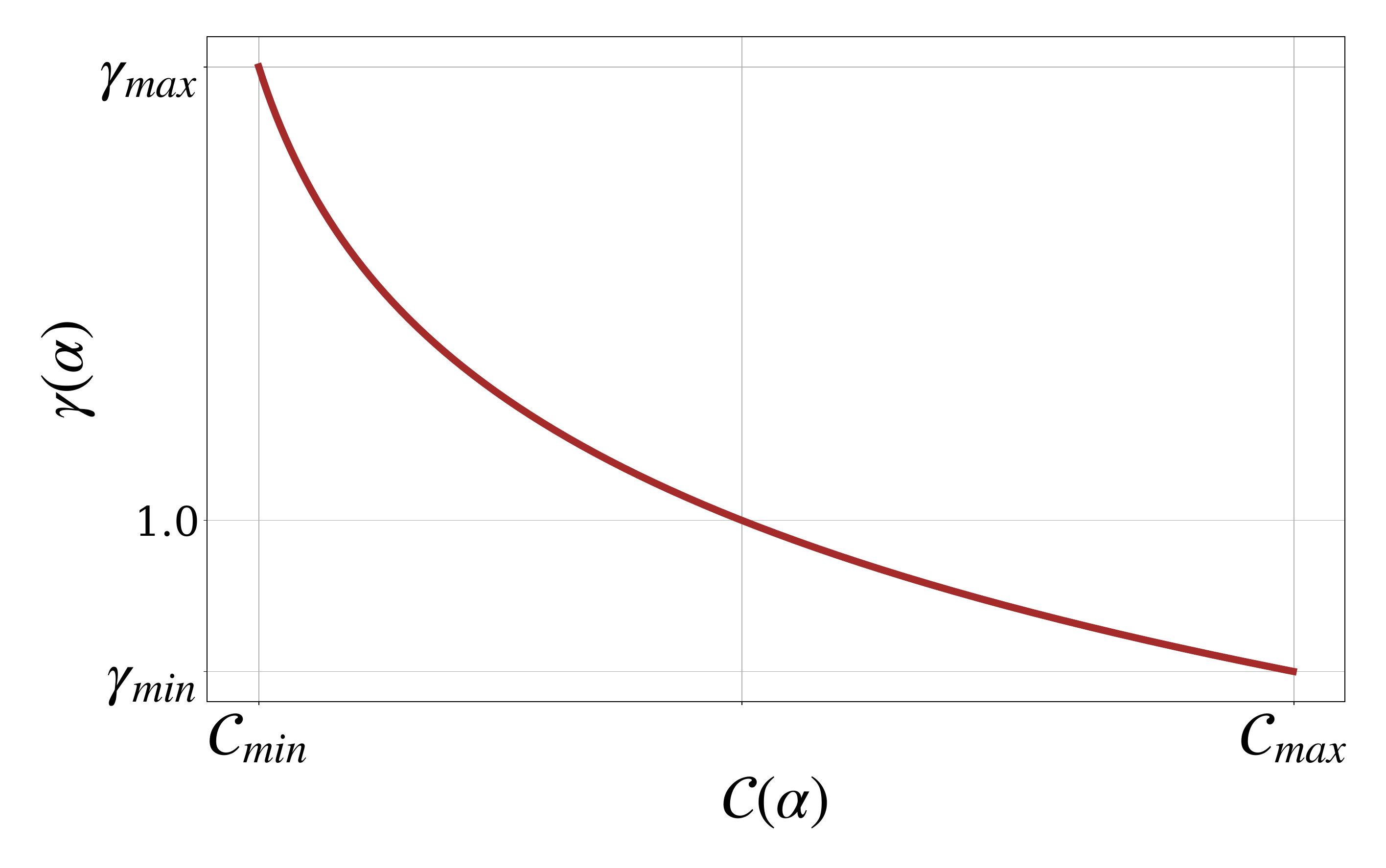}
        \vskip -0.1in
        \caption{}
        \label{fig:gamma_graph}
    \end{subfigure}
    \vspace{-0.2cm}
    \caption{(a) Plots of LRs by CaLR with varying the decay ratio of $\gamma(\alpha)$. CaLR sets a small decay ratio~(\ie,~a large LR) for high-complexity networks, and vice versa. (b) Visualization of the decay ratio $\gamma(\alpha)$ based on the complexity score $C(\alpha)$.}
    \label{fig:scaling_factor}
    \vspace{-0.3cm}
  \end{figure}

  \subsubsection{Complexity-aware LR Scheduler~(CaLR)} 

We define the complexity score of a subnet as the number of parameters to tune\footnote{Although we can exploit FLOPs as a complexity score, this requires more computations, compared to counting the number of parameters. In addition, we have empirically found that the number of parameters provides better results in terms of the Kendall’s Tau~(see the supplement).}. That is, the subnets of high complexities have more network parameters to learn, compared to those of low complexities, suggesting that the high-complexity subnets require more training steps for accurate ranking~(Figs.~\ref{fig:teaser}(a) and~\ref{fig:observation}(a)). Instead of directly training the subnets with more steps, which is computationally expensive, we propose to adjust a decay ratio of an LR scheduler adaptive to the complexity of subnets. Specifically, we decay LR values slowly for the subnets of high complexities, thereby anticipating more extensive training, while rapidly decaying them for the ones of low complexities~(Fig.~\ref{fig:scaling_factor}(a)).

Concretely, we formulate our CaLR as a polynomial function with a decay ratio~$\gamma(\alpha)$ as follows:

\begin{equation}
  \eta^{t} = \eta^{0} \cdot \left( 1 - \frac{t}{T} \right)^{\gamma(\alpha)},
  \label{eq:complexity_aware_scheduler}
\end{equation}
where $\eta^t$ is a LR at the $t$-th step, and $T$ is the total number of training steps. We lower the LR of high-complexity subnets slowly (\ie,~$\gamma_{\min}<\gamma(\alpha)<1$), while decaying it faster for low-complexity ones~(\ie,~$1<\gamma(\alpha)<\gamma_{\max}$)~(Fig.~\ref{fig:scaling_factor}(a)), where $\gamma_{\min}$ and $\gamma_{\max}$ are the minimum and maximum decay ratio of CaLR, respectively, which we set as hyperparameters~(see the supplement). To achieve this, we define the decay ratio~$\gamma(\alpha)$, which is inversely proportional to the complexity of the subnet, as follows:
\begin{equation}
  \gamma(\alpha) = \omega \log(\mathcal{C}(\alpha)) + \tau,
  \label{eq:complexity_decay_ratio}
\end{equation}
where $\mathcal{C}(\alpha)$ is a complexity score of a subnet $\alpha$, defined as the number of parameters for the subnet, and $\omega$ and $\tau$ are  coefficients for the affine transformation:
  \begin{flalign}
    \omega &= -\frac{\gamma_{\max} - \gamma_{\min}}{\log(\mathcal{C}_{\max}) - \log(\mathcal{C}_{\min})}, \\ 
    \tau &= \gamma_{\min} - \omega  \log(\mathcal{C}_{\max}).  	   
  \end{flalign}
We denote by $\mathcal{C}_{\min}$ and $\mathcal{C}_{\max}$ the minimum and maximum complexity scores of subnets in a search space, respectively. To compute the decay ratio~$\gamma(\alpha)$ in Eq.~\eqref{eq:complexity_decay_ratio}, the complexity score is fed into the logarithmic function~(Fig.~\ref{fig:scaling_factor}(b)), which is a key to differentiating the LR decay for the subnets of different complexities. We then apply an affine transform with the coefficients of $\omega$ and $\tau$ to calibrate the score of~$\log(\mathcal{C}(\alpha))$ within the range of $\gamma(\alpha)$. Note that the logarithmic function enables the subnet with a complexity score close to the medium (around the center point between $\mathcal{C}_{\min}$ and $\mathcal{C}_{\max}$) to be trained with a linearly decaying LR,~\ie,~$\gamma(\alpha)=1$~(Fig.~\ref{fig:scaling_factor}(b)).

Adjusting LRs based on the complexities of subnets by CaLR in Eq.~\eqref{eq:complexity_aware_scheduler} provides higher LRs for the subnets of high complexities, and vice versa. Considering that high-complexity subnets have lots of parameters to tune, CaLR has an effect of exploring the parameter space more thoroughly for the high-complexity subnets, mitigating the unfairness problem~(Fig.~\ref{fig:observation}(b)).

\subsubsection{Momentum Separation~(MS)}
\label{subsubsec:MS}  

\begin{figure}
  \captionsetup{font={small}}
  \begin{center}
    \begin{subfigure}{\linewidth}
         \centering
         \includegraphics[width=0.95\linewidth]{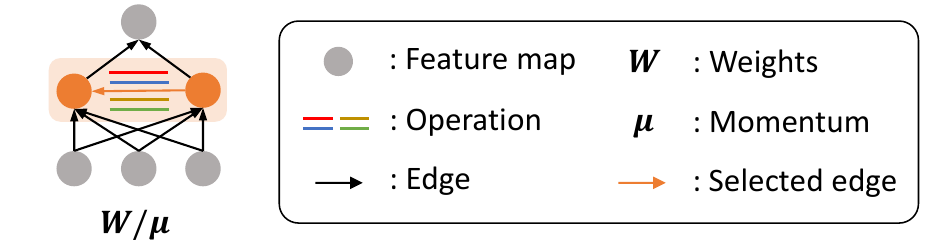}
         \caption{Supernet.}
         \label{fig:MS_a}
    \end{subfigure}
    \begin{subfigure}{\linewidth}
      \centering
      \includegraphics[width=0.95\linewidth]{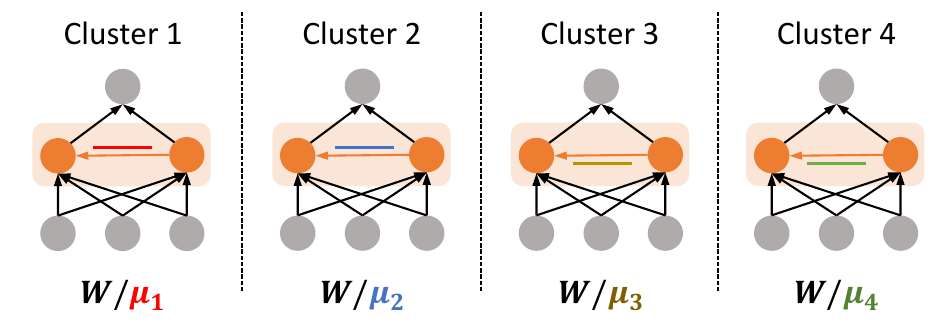}
      \caption{Momentum Separation~(MS).}
      \label{fig:MS_b}
 \end{subfigure}
\end{center}
\vspace{-0.5cm}
   \caption{(a)~The supernet shares weights and momentum for all subnets.~(b)~MS selects a single edge~(or layer) from the supernet and clusters the subnets according to operations for the edge. It then assigns a distinct momentum buffer for each cluster, while the weights are shared for all clusters.}
   \vspace{-0.3cm}
   \label{fig:MS}
 \end{figure}

Using a single optimizer for all subnets causes a noisy momentum problem. Since the sampled subnets vary at each training step, it is not enough to represent the gradients of individual subnets accurately by a single momentum buffer. It is highly likely that the gradients steer toward sub-optimal points~(Figs.~\ref{fig:teaser}(b) and~\ref{fig:observation}(c)). A straightforward solution is to use an individual momentum buffer for each subnet. This, however, requires substantial memory, due to huge amounts of subnets in a search space~(\eg,~$\sim7^{21}$ for MobileNet space~\cite{cai2018proxylessnas}) and the size of each momentum buffer, which is the same as the number of parameters in the subnet. MS instead clusters the subnets using the structural characteristics, and assigns a distinct momentum buffer to each cluster. The reason behind this is that subnets with similar structures provide similar gradients during training~\cite{kornblith2019similarity,peng2021pi}~(see the supplement for empirical verification). This suggests that the gradients of subnets from each cluster are likely to be consistent, alleviating the noisy momentum problem~(Fig.~\ref{fig:observation}(c)).

To this end, we consider that subnets are structurally similar, if they have the same operation at a specific edge~(or layer)~(Fig.~\ref{fig:MS})\footnote{In practice, we select the first layer for a macro search space~(\eg, MobileNet space~\cite{cai2018proxylessnas}), and a random edge for a micro search space~(\eg, NAS-Bench-201~\cite{dong2020bench}). See the supplement for more details.}, and cluster the subnets as follows:
\begin{equation}
  S_i = \{ \alpha\in\mathcal{A}~|~\text{op}(\alpha,e)=o_i \},~~~i\in[1,n],
  \label{eq:cluster}
\end{equation}
where $S_i$ is a cluster of subnets~$\alpha$ with the same operation $o_i$ at the chosen edge $e$, and $\text{op}(\alpha,e)$ represents an operation of the subnet $\alpha$ at edge $e$. Note that every edge in the supernet contains a set of candidate operations, $O=\{ o_1, o_2, ..., o_n \}$, where $n$ is the number of candidate operations, and the subnets adopt a single operation from the set of $O$ for every edge. For each cluster $S_i$, we assign a distinct momentum buffer~$\mu_i$. That is, if the sampled subnet $\alpha$ belongs to the cluster $S_i$, we update the momentum buffer $\mu_i$ at each training step, as follows:
\begin{equation}
  \mu_i^t = \beta \cdot \mu_i^{t-1} + g^t,
  \label{eq:momentum}
\end{equation}
where $\beta$ is a coefficient for moving average, and $g^t$ is the gradient w.r.t. a task loss at the $t$-th step. Note that MS separates the momentum update process only, while using shared weights across all subnets, suggesting that the overall search time remains the same. Note also that the memory overheads for the momentum buffer are negligible, compared to those for forward and backward passes during supernet training~\cite{gao2020estimating}. MS is also complementary to CaLR, and thus exploiting both CaLR and MS can further boost the ranking consistency~(Fig.~\ref{fig:observation}(d)).

\section{Experiments}
 \label{sec:exp}
 In this section, we first describe our experimental settings~(Sec.~\ref{subsec:exp_setup}). We then demonstrate the effectiveness of our approach through extensive experiments, including quantitative results on various search spaces and datasets~(Sec.~\ref{subsec:results}), and in-depth analyses on our method~(Sec.~\ref{subsec:Discussion}).

 \subsection{Experimental settings}
 \label{subsec:exp_setup}

  We validate our approach on widely used search spaces, including the MobileNet space~\cite{cai2018proxylessnas} and NAS-Bench-201~\cite{dong2020bench}. More details including the search spaces and hyperparameters are provided in the supplement.
  \paragraph{MobileNet.} 
  The MobileNet space~\cite{cai2018proxylessnas} consists of subnets with 21 mobile inverted bottleneck convolution~(MBConv) layers~\cite{sandler2018mobilenetv2}. Each MBConv layer has various parameters to determine, including the kernel sizes of~(3,~5,~7) and expansion ratios of~(3,~6). All layers except for the downsampling layers use a skip connection as an additional candidate operator, resulting in $\sim7^{21}$ size of the search space.

  We follow the standard settings~\cite{guo2020single,chu2021fairnas,ha2021sumnas,lu2023pa,oh2024efficient} for evaluating the search performance, consisting of supernet training, evolutionary search, and retraining. We train the supernet on ImageNet~\cite{deng2009imagenet} using the SGD optimizer with a momentum coefficient of 0.9 and a weight decay of 4e-5. We set an initial LR as 0.045 and decay the LR with our CaLR. We apply our dynamic supernet training to three NAS approaches, including one-shot NAS methods~(\ie, SPOS~\cite{guo2020single} and FairNAS~\cite{chu2021fairnas}) and a few-shot NAS method~(\ie, FSNAS~\cite{zhao2021few}). We use an evolutionary search~\cite{goldberg1991comparative} to discover the best subnet. We retrain the retrieved subnet to obtain the final accuracy, using the same setting of~\cite{lin2021zen,li2023zico,lee2024assembling}.

  \vspace{-0.35cm}
  \paragraph{NAS-Bench-201.} 

  NAS-Bench-201~\cite{dong2020bench} is a cell-based search space, which provides the stand-alone accuracies of each subnet on CIFAR-10, CIFAR-100~\cite{krizhevsky2009cifar} and ImageNet-16-120~\cite{chrabaszcz2017downsampled}. The cell structure of NAS-Bench-201 can be represented by a directed acyclic graph~(DAG)~with 4 nodes and 6 edges. Each edge in the DAG is characterized by one of the five candidate operations, including zeroize, skip-connect, 1$\times$1 and 3$\times$3 convolutions, and average pooling, leading to total $5^6$ subnets in the search space.
  
  We follow the supernet training scheme provided by the benchmark~\cite{dong2020bench}. Specifically, we use the SGD optimizer with a momentum coefficient of 0.9 and a weight decay of 5e-4. We use our CaLR as an LR scheduler with an initial LR of 0.025. We apply our method to SPOS~\cite{guo2020single}, FairNAS~\cite{chu2021fairnas} and FSNAS~\cite{zhao2021few} as in the MobileNet space. Following~\cite{ha2021sumnas}, we evaluate all subnets within the supernet to find the best one, and report the search performance in terms of the top-1 test accuracy of the retrieved subnet. To measure the ranking consistency of a supernet, we compute the Kendall's Tau~\cite{kendall1948rank} with a small set of subnets, where they are sampled following the strategy in~\cite{ha2021sumnas}, instead of using the whole subnets in the search space, which helps minimize noise in the ranking list~\cite{ha2021sumnas}. Specifically, we evenly split entire subnets into 400 chunks based on stand-alone accuracies, and sample the best-performing one in each chunk. We then compute the Kendall's Tau for the sampled subnets, using the rankings estimated by NAS methods and the stand-alone accuracies. For all experiments on NAS-Bench-201, we report the average and standard deviation over 3 runs.

  \begin{table}[t]
    \captionsetup{font={small}}
    \caption{Quantitative comparison of the search performance on ImageNet~\cite{deng2009imagenet} in the MobileNet space~\cite{cai2018proxylessnas}. We report the top-1 validation accuracy, together with the number of parameters and FLOPs for the retrieved subnets. We also report the peak memory usage, and the GPU hours for training supernets, computed with 4 A5000 GPUs.}
    \label{tab:imagenet}
    \vspace{-0.7cm}
    \small
    \setlength{\tabcolsep}{0.3em}
    \begin{center}
    \adjustbox{max width=\columnwidth}{
      \begin{tabular}{lcccccc}
        \toprule
        \multicolumn{1}{c}{Methods} & \begin{tabular}[c]{@{}c@{}}Params.\\ (M)\end{tabular} & \begin{tabular}[c]{@{}c@{}}FLOPs\\ (M)\end{tabular} & \begin{tabular}[c]{@{}c@{}}Top-1\\ (\%)\end{tabular} & \begin{tabular}[c]{@{}c@{}}Peak Memory\\ (MB)\end{tabular} & \begin{tabular}[c]{@{}c@{}}GPU Hours\\ (hours)\end{tabular} \\
        \midrule
        SPOS-S~\cite{guo2020single} & 4.1 & 330 & 75.6 & \multirow{3}{*}{54283} & \multirow{3}{*}{157} \\
        SPOS-M & 4.3 & 387 & 76.2 &  &  \\
        SPOS-L & 4.5 & 471 & 76.6 &  &  \\
        \rowcolor{Gray}
        SPOS-S + Ours & 3.9 & 329 & \textbf{75.9} &  & \\
        \rowcolor{Gray}
        SPOS-M + Ours & 4.5 & 396 & \textbf{76.7} &  &  \\
        \rowcolor{Gray}
        SPOS-L + Ours & 4.7 & 459 & \textbf{77.1} &  \multirow{-3}{*}{54579} & \multirow{-3}{*}{159} \\
        \midrule
        FairNAS-S~\cite{chu2021fairnas} & 4.1 & 330 & 75.7 & \multirow{3}{*}{54400} & \multirow{3}{*}{364} \\
        FairNAS-M & 4.3 & 394 & 76.3 &  &  \\
        FairNAS-L & 4.7 & 472 & 76.7 &  &  \\
        \rowcolor{Gray}
        FairNAS-S + Ours & 4.1 & 326 & \textbf{75.8} &  &  \\
        \rowcolor{Gray}
        FairNAS-M + Ours & 4.5 & 399 & \textbf{76.6} &  &  \\
        \rowcolor{Gray}
        FairNAS-L + Ours & 4.7 & 471 & \textbf{77.0} &  \multirow{-3}{*}{54612} & \multirow{-3}{*}{369}  \\
        \midrule
        FSNAS-S~\cite{zhao2021few} & 4.0 & 330 & 75.8 & \multirow{3}{*}{54537} & \multirow{3}{*}{740} \\
        FSNAS-M & 4.3 & 399 & 76.4 &  &  \\
        FSNAS-L & 4.7 & 464 & 76.8 &  &  \\
        \rowcolor{Gray}
        FSNAS-S + Ours & 4.0 & 324 & \textbf{76.0} &  &  \\
        \rowcolor{Gray}
        FSNAS-M + Ours & 4.4 & 398 & \textbf{76.6} &  &  \\
        \rowcolor{Gray}
        FSNAS-L + Ours & 4.5 & 467 & \textbf{77.2} &  \multirow{-3}{*}{54724} & \multirow{-3}{*}{744}  \\
        \bottomrule
  \end{tabular}
    }
      
  \vspace{-0.4cm}

    \end{center}
  
    \end{table}

  \subsection{Results}
 \label{subsec:results}
 We evaluate our method on various search spaces, including the MobileNet space~\cite{cai2018proxylessnas} and NAS-Bench-201 space~\cite{dong2020bench}. More quantitative results, including comparisons to the state of the art and evaluations on additional search spaces, are provided in the supplement.
 
 \vspace{-0.35cm}
 \paragraph{MobileNet.}
 We compare in Table~\ref{tab:imagenet} the performance of subnets retrieved by our method and the baselines on the MobileNet search space, in terms of the top-1 validation accuracy on ImageNet, together with the number of parameters and FLOPs of the searched subnets. We also report the peak memory usage, and the GPU hours for training supernets. We apply our dynamic training technique to SPOS~\cite{guo2020single}, FairNAS~\cite{chu2021fairnas} and FSNAS~\cite{zhao2021few}, where we denote by SPOS + Ours, FairNAS + Ours and FSNAS + Ours, respectively. We search the subnets with various FLOPs constraints including 330M~(small, S), 400M~(medium, M), and 475M~(large, L). We summarize the results of Table~\ref{tab:imagenet} in threefold: (1) Applying our dynamic training technique improves the performance of three baselines~(SPOS, FairNAS, and FSNAS), regardless of FLOPs constraints. This demonstrates the effectiveness of our dynamic supernet training scheme, compared to the static one. Note that each baseline exploits distinct supernet training strategy,~\eg, sampling a single~\cite{guo2020single} or multiple~\cite{chu2021fairnas} subnets at each training iteration, or using multiple sub-supernets~\cite{zhao2021few}. This suggests that our method can be applied in a plug-and-play manner across diverse supernet training algorithms. (2) The performance gain of our method is generally more significant for the subnets with higher complexities. This is because, the subnets with higher complexities are more likely to suffer from the unfairness problem, resulting in poor ranking consistency, and our method can alleviate the problem by providing a more balanced training for each subnet. (3) Additional memory costs and time consumptions for our method are negligible, compared to those for the baselines. The increase in memory comes from the additional momentum buffers in MS. It is negligible, compared with the overall memory usage for forward and backward passes~\cite{gao2020estimating}.  

 \begin{table*}[t]
  \centering
  \small 
  \setlength{\tabcolsep}{3pt}
  \caption{Quantitative comparison of different supernet training methods on CIFAR-10, CIFAR-100~\cite{krizhevsky2009cifar}, and ImageNet16-120~\cite{deng2009imagenet} datasets in NAS-Bench-201~\cite{deng2009imagenet}. We report the Kendall's Tau, along with the top-1 accuracy (Top-1 Acc.) for each method. We also report the peak memory usage, and the GPU hours for training supernets on CIFAR-10, computed with a single RTX 2080Ti. The results include the average and standard deviations for 3 runs.}
  \vspace{-0.2cm}
  \resizebox{\textwidth}{!}{
  \begin{tabular}{lcccccccc}
    \toprule
    \multicolumn{1}{c}{\multirow{2}{*}[-2pt]{\centering Methods}}  & \multicolumn{2}{c}{CIFAR-10} & \multicolumn{2}{c}{CIFAR-100} & \multicolumn{2}{c}{ImageNet16-120} & {\multirow{2}{*}[-2pt]{\shortstack[c]{Peak Memory \\  (MB) }}} & {\multirow{2}{*}[-2pt]{\shortstack[c]{GPU Hours \\  (hours) }}} \\
    \cmidrule(r){2-3} \cmidrule(r){4-5} \cmidrule(r){6-7}
    & Kendall's Tau & Top-1 Acc.   & Kendall's Tau & Top-1 Acc. & Kendall's Tau & Top-1 Acc. \\
    \midrule
    SPOS~\cite{guo2020single} & 0.751 $\pm$ 0.008 & 93.12 $\pm$ 0.03 & 0.787 $\pm$ 0.011 & 68.16 $\pm$ 0.35 & 0.718 $\pm$ 0.009 & 42.65 $\pm$ 0.33 & 1875 & 2.67 \\
    \rowcolor{Gray}
    SPOS + Ours & \textbf{0.814} $\pm$ 0.007 & \textbf{93.50} $\pm$ 0.33 & \textbf{0.81} $\pm$ 0.024 & \textbf{69.72} $\pm$ 0.41 & \textbf{0.743} $\pm$ 0.011 & \textbf{43.30} $\pm$ 0.70 & 1901 & 2.69 \\
    \midrule
    FairNAS~\cite{chu2021fairnas} & 0.766 $\pm$ 0.015 & 92.13 $\pm$ 0.18 & 0.79 $\pm$ 0.009 & 67.72 $\pm$ 0.40 & 0.699 $\pm$ 0.008 & 39.81 $\pm$ 0.42 & 1877 & 5.13 \\
    \rowcolor{Gray}
    FairNAS + Ours & \textbf{0.828} $\pm$ 0.020 & \textbf{93.52} $\pm$ 0.50 & \textbf{0.857} $\pm$ 0.015 & \textbf{70.91} $\pm$ 1.35 & \textbf{0.776} $\pm$ 0.016 & \textbf{44.63} $\pm$ 0.37 & 1903 & 5.15 \\
    \midrule
    FSNAS~\cite{zhao2021few} & 0.729 $\pm$ 0.019 & 93.43 $\pm$ 0.24 & 0.757 $\pm$ 0.013 & 69.49 $\pm$ 0.08 & 0.688 $\pm$ 0.018 & 42.97 $\pm$ 0.52 & 1901 & 16.04 \\
    \rowcolor{Gray}
    FSNAS + Ours & \textbf{0.767} $\pm$ 0.010 & \textbf{93.63} $\pm$ 0.21 & \textbf{0.774} $\pm$ 0.028 & \textbf{71.05} $\pm$ 0.17 & \textbf{0.728} $\pm$ 0.017 & \textbf{44.68} $\pm$ 0.06 & 1925 & 16.12 \\
    \midrule
    Optimal & - & 94.37 & - & 73.51 & - & 47.31 & - & - \\ 
    \bottomrule
\end{tabular}
}
\vspace{-0.2cm}

  \label{tab:nb201}
\end{table*}

  \vspace{-0.35cm}
   \paragraph{NAS-Bench-201.}
We show in Table~\ref{tab:nb201} the search performance in the NAS-Bench-201 space, in terms of ranking consistencies and top-1 accuracies. We can see that the three baselines~\cite{guo2020single,chu2021fairnas,zhao2021few} coupled with our dynamic supernet training method provide better search performance consistently with negligible additional search cost. This confirms once again that our dynamic supernet training technique can be an effective alternative to the static one.

  \subsection{Discussion}
  \label{subsec:Discussion} 
We provide an ablation study and in-depth analyses on each component of our method, including CaLR and MS. For more discussions, please refer to the supplement.
    
    \begin{table}[t]
          \captionsetup{font={small}}
          \small
         
          \caption{Quantitative comparison of the ranking consistency on CIFAR-10, CIFAR-100~\cite{krizhevsky2009cifar}, and ImageNet16-120~\cite{deng2009imagenet} datasets in NAS-Bench-201~\cite{deng2009imagenet}. We apply CaLR and MS to SPOS~\cite{guo2020single}, FairNAS~\cite{chu2021fairnas}, and FSNAS~\cite{zhao2021few}, and report Kendall’s Tau / Accuracy.}

        \label{tab:ablation}
        \vspace{-0.2cm}
          \centering
          \setlength{\tabcolsep}{2pt}
        \resizebox{\columnwidth}{!}{
          \begin{tabular}{cccccc}
            \toprule 
            Baselines    & CaLR & MS  & CIFAR-10          & CIFAR-100         & ImageNet16-120 \\
            \midrule
            \multirow{4}{*}{SPOS [12]}   & -    & -   & $0.751 / 93.12$     & $0.787 / 68.16$     & $0.718 / 42.65$ \\
                                         & \checkmark & -   & $0.805 / 93.3$     & $0.803 / 68.94$     & $0.733 / 42.83$ \\
                                         & -    & \checkmark & $0.772 / 93.45$     & $0.807 / 69.11$     & $0.724 / 43.04$ \\
                                         & \checkmark & \checkmark & $\mathbf{0.814 / 93.5}$ & $\mathbf{0.81 / 69.72}$ & $\mathbf{0.743 / 43.3}$ \\
            \midrule
            \multirow{4}{*}{FairNAS [7]} & -    & -   & $0.766 / 92.13$     & $0.790 / 67.72$     & $0.699 / 39.81$ \\
                                         & \checkmark & -   & $0.819 / 93.21$     & $0.832 / 69.29$     & $0.74 / 43.13$ \\
                                         & -    & \checkmark & $0.784 / 92.89$     & $0.815 / 68.93$     & $0.725 / 42.89$ \\
                                         & \checkmark & \checkmark & $\mathbf{0.828 / 93.52}$ & $\mathbf{0.857 / 70.91}$ & $\mathbf{0.776 / 44.63}$ \\
            \midrule
            \multirow{4}{*}{FSNAS [35]}  & -    & -   & $0.729 / 93.43$     & $0.757 / 69.49$     & $0.688 / 42.97$ \\
                                         & \checkmark & -   & $0.749 / 93.57$     & $0.764 / 70.52$     & $0.71 / 43.32$ \\
                                         & -    & \checkmark & $0.750 / 93.52$     & $0.769 / 69.97$     & $0.705 / 43.51$ \\
                                         & \checkmark & \checkmark & $\mathbf{0.767 / 93.63}$ & $\mathbf{0.774 / 71.05}$ & $\mathbf{0.728 / 44.68}$ \\
            \bottomrule
        \end{tabular}
        }
        \vspace{-0.4cm}
          
      \end{table}
      
      \vspace{-0.35cm}
      \paragraph{Ablation study.} We report in Table~\ref{tab:ablation} the ablation study on each component of our method. We apply CaLR and MS to SPOS~\cite{guo2020single}, FairNAS~\cite{chu2021fairnas} and FSNAS~\cite{zhao2021few}, and measure the Kendall's Tau and top-1 accuracy. We can see that applying either CaLR or MS improves the ranking consistency and search performance of the baselines. This implies that the static training scheme suffers from the unfairness and noisy momentum problems, regardless of the type of NAS methods, which can be alleviated by our dynamic training. We also observe that CaLR and MS can be complementary to each other, and applying both CaLR and MS further improves the ranking consistency.

      \vspace{-0.35cm}
      \paragraph{Unfairness Problem.} To see how the complexities of subnets affect a training process, we introduce two metrics, Complexity Bias~(CB) and Complexity-Convergence Correlation~($C^3$). First, CB is designed to detect the bias caused by the complexities of subnets. Specifically, it computes the ratio of subnet pairs, where the subnet predicted in low rank has a higher complexity, but it is supposed to provide better performance:
      \vspace{-0.2cm}
      \begin{equation}
        \text{CB}= \frac{\Sigma_{(\alpha,\beta)\in \mathcal{S}}\mathbb{I}(\alpha,\beta) \mathbb{J}(\alpha,\beta)}{\Sigma_{(\alpha,\beta)\in \mathcal{S}}\mathbb{I}(\alpha,\beta)},
        \vspace{-0.2cm}
      \end{equation}
      where $\mathcal{S}$ is a set of all possible pairs of subnets, and $(\alpha, \beta)$ is a pair of subnets within the set. The indicator function~$\mathbb{I}(\alpha,\beta)$ returns 1, when the predicted ranking of the pair $(\alpha,\beta)$ misaligns with the ground truth, and 0 otherwise. The function~$\mathbb{J}(\alpha,\beta)$ is similarly defined but returns~1, when the subnet in the pair predicted in low rank has a higher complexity. Second, \textit{$C^3$} identifies the relationship between the complexity of subnets and the convergence ratio:
      \vspace{-0.2cm}
      \begin{equation}
        C^3= \tau(\mathcal{C}(\mathcal{A}),\text{CR}(\mathcal{A})),
        \vspace{-0.2cm}
      \end{equation}
      where $\tau(\cdot)$ outputs the Kendall's Tau between inputs. $\mathcal{C}(\mathcal{A})$ and $\text{CR}(\mathcal{A})$ are the set of complexity scores and the convergence ratio of subnets in search space $\mathcal{A}$, respectively. The convergence ratio is defined as the accuracy estimated from the supernet over the stand-alone accuracy. Note that CB and $C^3$ approach to 0.5 and 0, respectively, as the ranking predictions of the supernet become less related to the subnet complexity. 

      \begin{table}[t]
        \captionsetup{font={small}}
        \caption{Quantitative comparisons of \textit{CB} and \textit{$C^3$} on CIFAR-10~\cite{krizhevsky2009cifar} in the NAS-Bench-201 space~\cite{dong2020bench}. Note that we sample the top 30\% of subnets~(4688 out of $5^6$) based on the stand-alone accuracies to compute CB and \textit{$C^3$}, and we set $k$ to $\mathcal{C}_\text{min}+\mathcal{C}_\text{max}$. See text for more details.}
        \label{tab:unfairness}
        \vspace{-0.2cm}
        \centering
        \small
        \resizebox{0.95\linewidth}{!}{
        \begin{tabular}{ccccc}
          \toprule
          \begin{tabular}[c]{@{}c@{}}Baseline\end{tabular} & CaLR & $\gamma(\alpha)\propto$ & CB & \textit{$C^3$} \\
          \midrule
          \multirow{3}{*}{SPOS~\cite{guo2020single}} & - & - & 0.76 & -0.19 \\
                                & $\checkmark$ & $-\log(k-\mathcal{C}(\alpha))$ & 0.79 & -0.22\\
                                & $\checkmark$ & $-\log(\mathcal{C}(\alpha))$ & \textbf{0.63} &\textbf{-0.09} \\
        \bottomrule   
        \end{tabular}
        }
        
        \vspace{-5mm}
      \end{table}

      We show in Table~\ref{tab:unfairness} the results of CB and \textit{$C^3$} on CIFAR-10 in NAS-Bench-201~\cite{dong2020bench}. We can see that the supernet trained using SPOS~\cite{guo2020single} provides CB of 0.76, which indicates that the supernet mis-ranks the subnets mainly due to the complexities. At the same time, the supernet shows a negative correlation between the convergence ratio and the subnet complexity~(\ie,~\textit{$C^3$} of -0.19). This suggests that high-complexity subnets tend to be insufficiently trained compared to low-complexity ones, leading to the unfairness problem. From the first and third rows, we can see that applying CaLR to SPOS boosts the performance in terms of CB and $C^3$. This shows that CaLR provides more adequate training for the subnets according to their complexities, which is a key factor in alleviating the unfairness problem. Note that the LR decay ratio of CaLR is inversely proportional to the complexity score~(\ie,~$\gamma(\alpha)\propto-\log(\mathcal{C}(\alpha))$). To further validate the effectiveness of CaLR, we apply it with a decaying strategy of an opposite effect, where the decay ratio is proportional to the complexity score of the subnet~(\ie,~$\gamma(\alpha)\propto-\log(k-\mathcal{C}(\alpha))$, where $k=\mathcal{C}_\text{min}+\mathcal{C}_\text{max}$). We can see from the second row that this variant rather degrades the performance in terms of CB and $C^3$, compared to the baseline, demonstrating that our LR strategy is effective in mitigating the unfairness problem.

      \begin{table}[t]
         \captionsetup{font={small}}
         \caption{Quantitative results of the ranking consistency on CIFAR-10~\cite{krizhevsky2009cifar} in the NAS-Bench-201 space~\cite{dong2020bench} with different subnet clustering algorithms.}
         \label{tab:MS}
         \vspace{-0.2cm}
         \centering
         \small

         \begin{tabular}{cccc}
           \toprule
           \begin{tabular}[c]{@{}c@{}}Baseline\end{tabular} & MS & Clustering & \begin{tabular}[c]{@{}c@{}}Kendall's Tau\end{tabular} \\
           \midrule
           \multirow{3}{*}{SPOS~\cite{guo2020single}} & - & - & 0.751$\pm$0.008  \\
                                 & $\checkmark$ & Random & 0.750$\pm$0.019 \\
                                 & $\checkmark$ & Operation-based & \textbf{0.772}$\pm$0.007  \\
         \bottomrule   
         \end{tabular}
         
         
        \end{table}
        \begin{figure}[t]
      
         \captionsetup{font={small}}
         \centering
         \hspace*{\fill}
         \begin{subfigure}{0.48\linewidth}
             \centering
             \includegraphics[width=0.9\linewidth]{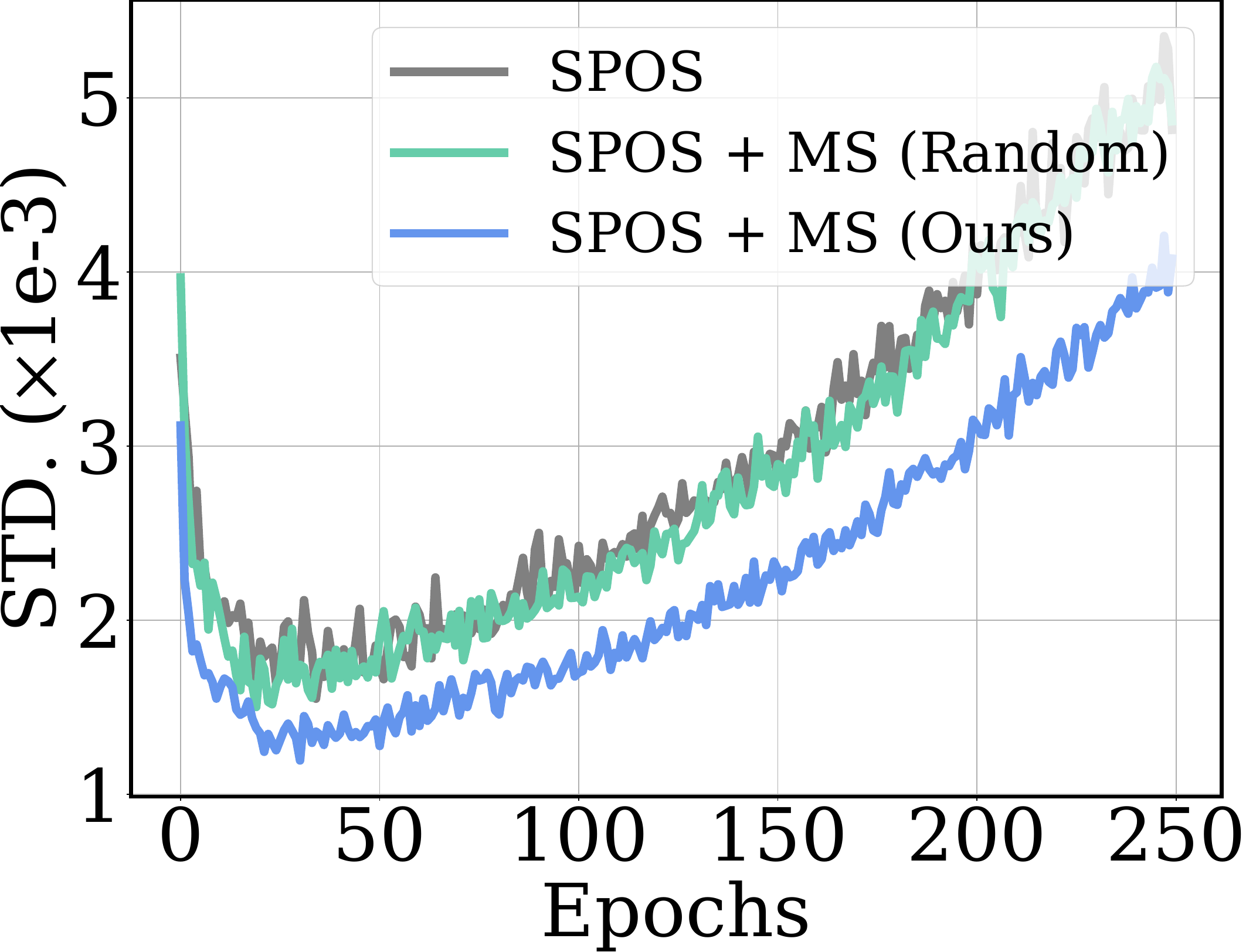}
             \caption{Gradients.}
             \label{fig:gradients}
         \end{subfigure}
         \hfill
         \begin{subfigure}{0.48\linewidth}
             \centering
             \includegraphics[width=0.9\linewidth]{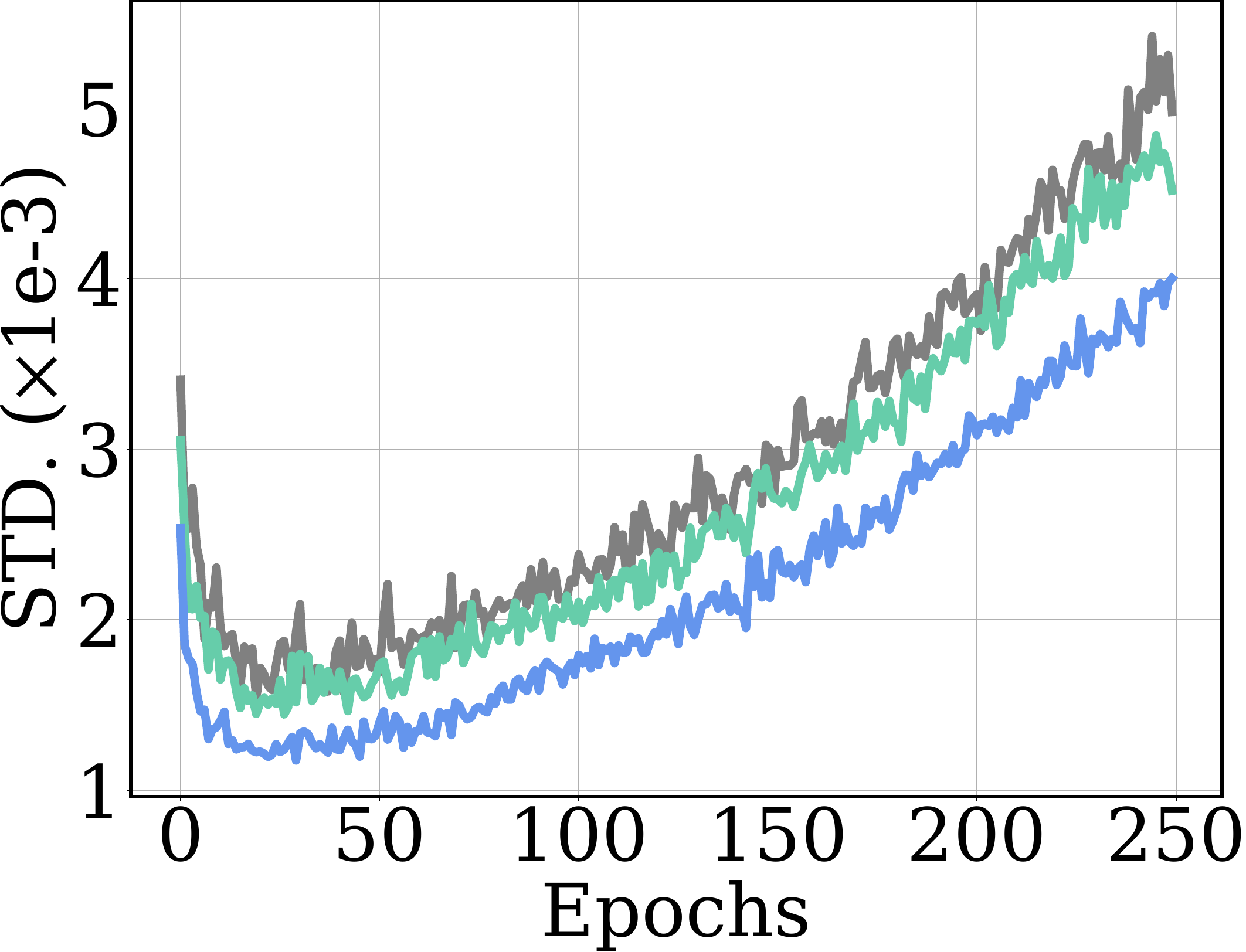}
             \caption{Momentums.}
             \label{fig:momentums}
         \end{subfigure}
         \hspace*{\fill}
         \vspace{-0.2cm}
         \captionof{figure}{Plots of standard deviations for gradients and momentums of the supernet on CIFAR-10~\cite{krizhevsky2009cifar} in NAS-Bench-201~\cite{dong2020bench}.}
         
         \label{fig:gsnr}
          \vspace{-0.5cm}
       \end{figure}

      \vspace{-0.35cm}
    \paragraph{Clustering algorithms.} We show in Table~\ref{tab:MS} the effectiveness of MS in terms of the ranking consistency on CIFAR-10 in NAS-Bench-201. We apply MS to SPOS~\cite{guo2020single} with various clustering strategies. For the random clustering, subnets are randomly divided into 5 clusters, and assign a distinct momentum for each cluster, while our approach groups the subnet based on the shared operations at a chosen edge. We can see from the first and the third rows that our approach enhances the performance of SPOS~\cite{guo2020single}. On the other hand, the random clustering does not provide a performance gain, and the randomness causes the large variance of the Kendall's Tau, suggesting that the MS is largely influenced by grouping techniques. We visualize in Fig.~\ref{fig:gsnr} gradient and momentum consistencies. Following~\cite{lu2023pa}, we compute the average standard deviations of gradients and momentums to measure the consistencies. We can see that applying our approach to SPOS reduces the standard deviations of gradients and momentums, demonstrating the effectiveness of our approach in handling the noisy momentum problem. Moreover, we observe that our clustering approach provides much lower standard deviations of gradients and momentums, compared to the random clustering.

    \begin{table}[t]
      \captionsetup{font={small}}
      \caption{Quantitative results of the ranking consistency on CIFAR-10~\cite{krizhevsky2009cifar} in NAS-Bench-201~\cite{dong2020bench} with different numbers of clusters. We also report the peak memory usage in supernet training.}
      \vspace{-0.3cm}
      \setlength{\tabcolsep}{3pt}
      \centering
      \small
      \adjustbox{max width=\linewidth}{
      \begin{tabular}{cccc}
        \toprule
        Baseline & \# of clusters & Kendall’s Tau & Peak Memory (MB) \\
        \midrule
        \multirow{4}{*}{SPOS~\cite{guo2020single}} & 5 & 0.81$\pm$0.024 & \textbf{1901} \\
        &25 & \textbf{0.813$\pm$0.017} & 2187 \\
        &125 & 0.781$\pm$0.022 & 2839 \\
        &625 & 0.752$\pm$0.027 & 6059 \\
        \bottomrule
        \end{tabular}
      }
        \label{tab:num_clusters}
        \vspace{-0.4cm}
      \end{table}

      \vspace{-0.5cm}
    \paragraph{Number of clusters.}
    We show in Table~\ref{tab:num_clusters} ablations on the number of clusters for MS on CIFAR-10 in NAS-Bench-201. MS randomly selects a single edge from the supernet, and clusters the subnets according to operations for the edge. That is, we use five clusters, since each edge provides five candidate operations in the search space. In this experiment, to vary the number of clusters, we select 2 edges ($5^2$ = 25 clusters), 3 edges ($5^3$ = 125 clusters), and 4 edges ($5^4$ = 625 clusters) for clustering in MS. We can observe from the results that increasing the number of clusters from 5 to 25 yields a slight improvement in ranking consistency at the expense of more memory footprints, suggesting that 5 clusters (\ie, selecting a single edge for clustering) is enough, in terms of the ranking consistency and memory, within our framework. We also see that choosing the number of clusters as 125 or 625 rather degrades the ranking consistency. This is because, with a substantial number of momentum clusters, each momentum buffer receives insufficient updates. The momentum buffer, storing the running mean of gradients to stabilize the training process, becomes less informative, which can negatively affect training supernets.

    \section{Limitations}
    \label{sec:app_limitation}
    Our framework can be applied to sampling-based NAS methods~\cite{guo2020single,hu2020angle,chen2021one,peng2020cream,wang2021alphanet, chu2021fairnas,ha2021sumnas}, but not to differentiable NAS methods~\cite{liu2018darts, chen2019progressive,chu2020darts, ye2022b,chu2020fairdarts}. This is because our framework considers the different characteristics of subnets sampled from the supernet, while differentiable NAS methods jointly optimize the supernet weights and operation importance coefficients, without explicitly considering the individual subnets. Despite this limitation, the growing interest in sampling-based methods over differentiable ones, due to their memory efficiency, highlights the importance of our work.

\section{Conclusion}
We have introduced a dynamic supernet training technique that considers the distinct characteristics of individual subnets for \textit{N}-shot NAS. Specifically, we have proposed CaLR and MS to alleviate the unfairness and noisy momentum problems, respectively. We have demonstrated that our method can improve the search performance of various NAS methods, without bells and whistles. We believe that our method can be a strong baseline for future research on NAS.

\section*{Acknowledgements}
This work was partly supported by IITP grant funded by the Korea government (MSIT) (No.RS-2022-00143524, Development of Fundamental Technology and Integrated Solution for Next-Generation Automatic Artificial Intelligence System, No.2022-0-00124, RS-2022-II220124, Development of Artificial Intelligence Technology for Self-Improving Competency-Aware Learning Capabilities) and the Yonsei Signature Research Cluster Program of 2024~(2024-22-0161).

{
    \small
    \bibliographystyle{ieeenat_fullname}
    \bibliography{main}
}

\clearpage
\newpage





%
\renewcommand{\theequation}{\roman{equation}} 
\setcounter{equation}{0} 

\renewcommand{\thefigure}{\Alph{figure}} 
\setcounter{figure}{0} 

\renewcommand{\thetable}{\Alph{table}} 
\setcounter{table}{0} 

\renewcommand{\thesection}{\Alph{section}} 
\setcounter{section}{0} 


\maketitlesupplementary

In the supplementary material, we provide detailed implementation descriptions~(Sec.~\ref{sec:app_implementation}), and additional results, including comparisons to the state-of-the-art methods, and experiments conducted on additional NAS spaces~(Sec.~\ref{sec:app_exp}). We also present discussions on the ablation studies and the design choices of our method~(Sec.~\ref{sec:app_discussion}). Furthermore, we explain extensions to various \textit{N}-shot NAS methods with detailed algorithmic explanations and pseudocode~(Sec.~\ref{sec:app_algorithm}). Lastly, we visualize the structures of the subnets searched with our method~(Sec.~\ref{sec:app_searched_subnets}).

\section{Implementation details}
  \label{sec:app_implementation}
  \paragraph{MobileNet.} 
  We follow the standard approach~\cite{guo2020single,chu2021fairnas} for supernet training, evolutionary search, and retraining the retrieved subnet. For training supernet on ImageNet~\cite{deng2009imagenet} in the MobileNet space~\cite{cai2018proxylessnas}, we use the SGD optimizer with a momentum coefficient of 0.9 and a weight decay of 4e-5. The batch size and initial LR are set as 512 and 0.045, respectively. We apply our dynamic supernet training to three baselines, including SPOS~\cite{guo2020single}, FairNAS~\cite{chu2021fairnas}, and FSNAS~\cite{zhao2021few}, where they are trained for 150, 75, and 100 epochs, respectively. Note that FSNAS utilizes multiple sub-supernets, and each sub-supernet is trained for 100 epochs. We set the maximum and minimum decay ratio of CaLR as $\gamma_\text{max} = 1/\gamma_\text{min} = \gamma'$, and perform a grid search to determine $\gamma' \in \{2,3,4\}$, and we determine $\gamma' = 3$ for all baselines. For layer selection in MS, we choose the first one of the searchable layers. We then perform the evolutionary search for 20 epochs, with a population number of 50 to find the best subnet. For retraining the retrieved subnets, we use the same setting of~\cite{lin2021zen,li2023zico}. We use 4 A5000 GPUs for training the supernet and 8 A5000 GPUs for retraining the searched subnets.

  \paragraph{NAS-Bench-201.} 
  We train the supernet with the SGD optimizer with a momentum coefficient of 0.9 and a weight decay of 5e-4. The batch size and initial LR are set as 64 and 0.025, respectively. Our framework is applied on SPOS~\cite{guo2020single}, FairNAS~\cite{chu2021fairnas}, and FSNAS~\cite{zhao2021few}, where they are trained for 250, 200, and 300 epochs, respectively. Similar to the MobileNet space, we perform the grid search to determine the maximum and minimum decay ratio of CaLR in the same candidate, and we set $\gamma' = 4$. We use a randomly selected edge for subnet clustering in MS. We use a single RTX 2080Ti for training the supernet.

  \begin{table}[t]
    \captionsetup{font={small}}
     \small
     \caption{Quantitative comparison of the search performance on CIFAR-10~\cite{krizhevsky2009cifar} in the NAS-Bench-201 space~\cite{dong2020bench}.}
     \label{tab:cifar}
    \centering
    \small
     \begin{tabular}{lc}
       \toprule
       \multicolumn{1}{c}{Methods} & Top-1 Acc. \\
     \midrule
     DARTS~\cite{liu2018darts} & $86.23\pm4.93$ \\
     GDAS~\cite{dong2019gdas} & $93.26\pm0.32$ \\
     NSAS~\cite{zhang2020overcoming} & $92.23\pm0.10$ \\
     Cream~\cite{peng2020cream} & $92.83\pm0.67$ \\
     SUMNAS~\cite{ha2021sumnas} & $93.09\pm0.12$ \\
     PA\&DA~\cite{lu2023pa} & $93.33\pm0.22$ \\
     \midrule
     SPOS~\cite{guo2020single} & $93.12\pm0.03$ \\
     \rowcolor{Gray}
     SPOS + Ours & $\textbf{93.50}\pm0.33$ \\
     \midrule
     FairNAS~\cite{chu2021fairnas} & $92.13\pm0.18$ \\
     \rowcolor{Gray}
     FairNAS + Ours & $\textbf{93.52}\pm0.50$ \\
     \midrule
     FSNAS~\cite{zhao2021few} & $93.43\pm0.24$ \\
     \rowcolor{Gray}
     FSNAS + Ours & $\textbf{93.63}\pm0.21$ \\
     \bottomrule
     \end{tabular}
  
  \end{table}
  
  \begin{figure}[t]
    \captionsetup{font={small}}
      \centering
      \includegraphics[width=0.8\linewidth]{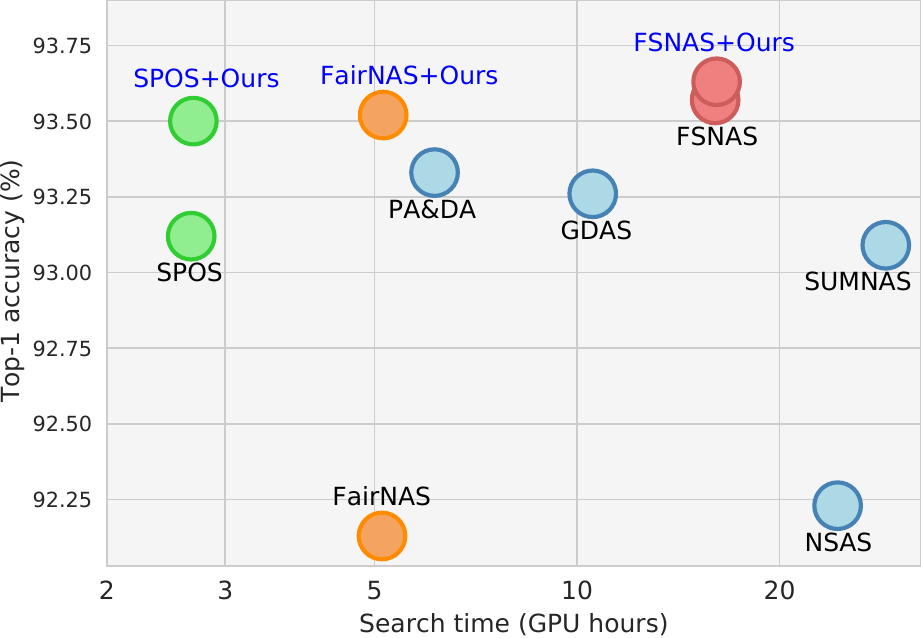}
      \captionof{figure}{Visual comparison of the top-1 accuracy and search time for the subnets retrieved from the NAS-Bench-201 space~\cite{dong2020bench} for CIFAR-10~\cite{krizhevsky2009cifar}.}
      \label{fig:search_time}
   \end{figure}

  \begin{table}[t]
    \setlength{\tabcolsep}{0.3em}
      \captionsetup{font={small}}
      \captionof{table}{Quantitative results of the top-1 accuracy on CIFAR-10 in the DARTS space.}
    \centering
    \small
  \label{tab:darts}
  \begin{tabular}{lc}
  \toprule
  \multicolumn{1}{c}{Methods}       & Top-1 Acc. (\%) \\ \midrule
  SPOS        & 96.1            \\ 
  SPOS + Ours & \textbf{96.3}            \\ \bottomrule
  \end{tabular}
    \end{table}

    \begin{table}[t]
    \captionsetup{font={small}}
    \setlength{\tabcolsep}{0.3em}
      \captionof{table}{Quantitative comparisons on ImageNet in the Autoformer space. We report the top-1 validation accuracy, together with the number of parameters limit and the number of parameters.}
    \centering
    \small
    \label{tab:autoformer}
    \begin{tabular}{lccc}
    \toprule
    \multicolumn{1}{c}{Methods}               & \# Params Limit & \# Params & Top-1 Acc. (\%) \\ \midrule
    Autoformer            & $\leq$6M        & 5.97M     & 74.5            \\ 
    Autoformer + Ours     & $\leq$6M        & 5.98M     & \textbf{74.8}            \\ \midrule
    Autoformer            & $\leq$7.5M      & 7.49M     & 76.0            \\ 
    Autoformer + Ours     & $\leq$7.5M      & 7.49M     & \textbf{76.3  }          \\ \midrule
    Autoformer            & $\leq$9M        & 8.91M     & 76.5            \\ 
    Autoformer + Ours     & $\leq$9M        & 8.96M     & \textbf{76.7}            \\ \bottomrule
    \end{tabular}
  \end{table}

\section{More experiments}
\label{sec:app_exp}
In this section, we provide a comparison to the state-of-the-art NAS methods, and results on additional NAS spaces, including DARTS~\cite{liu2018darts} and Autoformer spaces~\cite{chen2021autoformer}.
\subsection{Comparison to the state of the art}
We present in Table~\ref{tab:cifar} the quantitative comparison of our method and state-of-the-art NAS methods for NAS~\cite{liu2018darts,dong2019gdas,zhang2020overcoming,peng2020cream,ha2021sumnas,lu2023pa,guo2020single,chu2021fairnas,zhao2021few} on CIFAR-10~\cite{krizhevsky2009cifar} in the NAS-Bench-201 space~\cite{dong2020bench}. We can see that our method outperforms state-of-the-art methods in terms of top-1 accuracy. To further validate the efficiency of our method, we compare in Fig.~\ref{fig:search_time} the search performance and retrieval time. We can observe that ours provides better compromise between the search performance and search time compared to other methods. Moreover, our method can improve the performance of various NAS approaches, without a significant increase in the search time. For example, applying a dynamic supernet training to FairNAS~\cite{chu2021fairnas} boosts the top-1 accuracy of the searched subnets from 92.13\% to 93.52\%, with a negligible additional time, demonstrating the efficiency of our method.

\subsection{Results on additional spaces}
To show that our method is generally applicable to various NAS spaces, we provide other benchmark results, including DARTS~\cite{liu2018darts} and Autoformer spaces~\cite{chen2021autoformer}. The DARTS space is a cell-based search space, searching for normal and reduction cells, each consisting of 14 edges and 7 candidate operations. We train a supernet with SPOS~\cite{guo2020single} and SPOS with our dynamic supernet training on CIFAR-10. We follow the setting of~\cite{li2020random} for training a supernet and searching subnets. Specifically, the supernet is trained for 50 epochs, and the most promising subnet is selected by evaluating 1,000 subnets randomly chosen. We retrain the searched subnet with 200 epochs.

The Autoformer space~\cite{chen2021autoformer} is designed for building ViT architectures, with search parameters including Q-K-V dimension, embedding dimension, number of heads, MLP ratio, and network depth. We follow the experimental setting of~\cite{chen2021autoformer} for supernet training and subnet searching. We use the Supernet-tiny of the Autoformer space as our supernet, and train the supernet on ImageNet using the Autoformer algorithm~\cite{chen2021autoformer} with and without our approach. We then search the subnets using an evolutionary algorithm with various constraints on the number of parameters. Note that the Autoformer algorithm does not require a retraining step.

We show in Tables~\ref{tab:darts} and~\ref{tab:autoformer} quantitative comparisons of baseline NAS methods with and without our approach. We can see that our method can generally enhance these baselines in various search spaces. Note that these improvements come from marginal overheads, demonstrating that our method can be applied efficiently to various scenarios.

\begin{figure*}[t]
  \centering
  \footnotesize
  \small
  \begin{subfigure}[b]{0.25\textwidth}
      \centering
      \includegraphics[width=\textwidth]{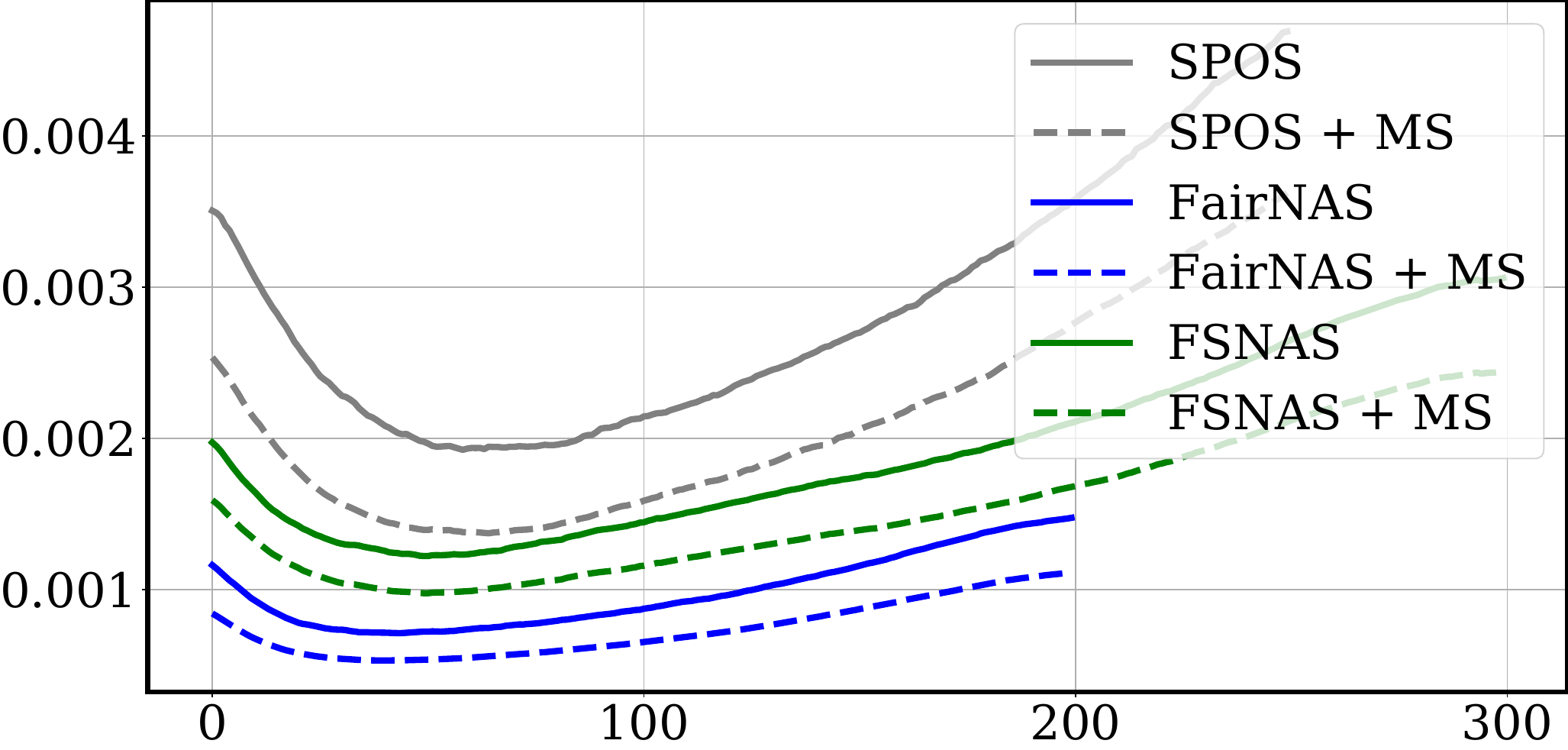} 
      \caption{CIFAR-10}
      \label{fig:subfig-a}
  \end{subfigure}
  \qquad
  \begin{subfigure}[b]{0.25\textwidth}
      \centering
      \includegraphics[width=\textwidth]{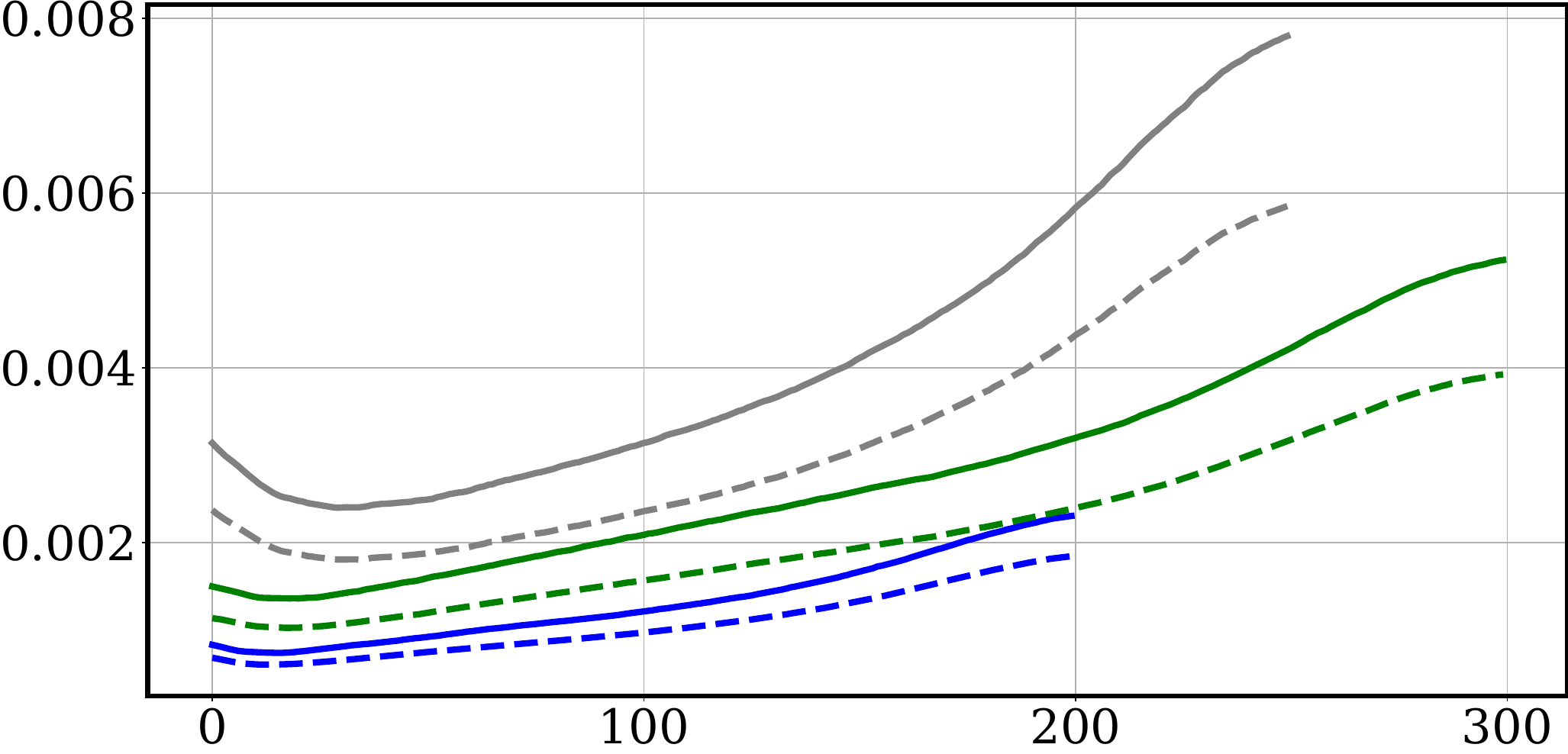} 
      \caption{CIFAR-100}
      \label{fig:subfig-b}
  \end{subfigure}
  \qquad
  \begin{subfigure}[b]{0.25\textwidth}
      \centering
      \includegraphics[width=\textwidth]{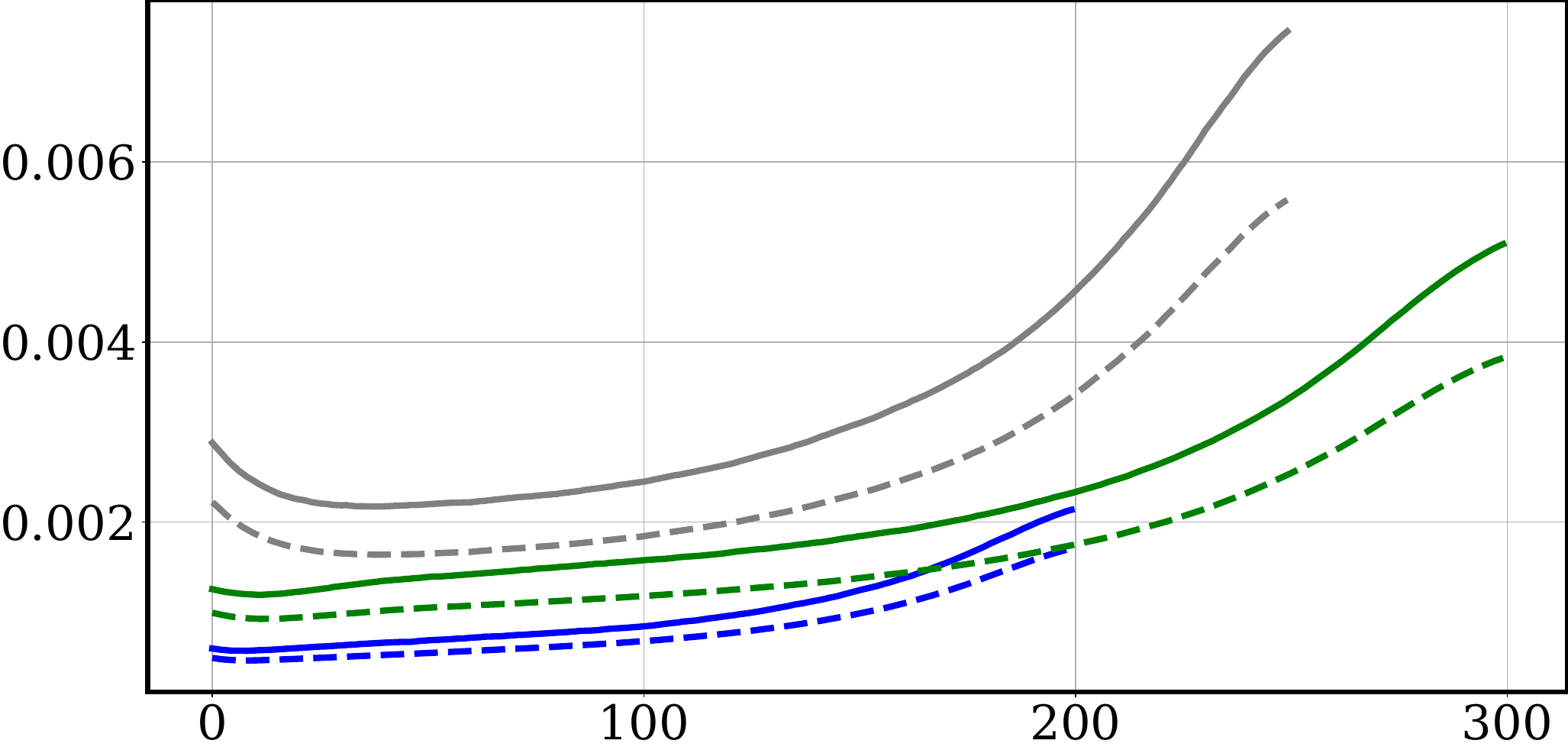} 
      \caption{ImageNet16-120}
      \label{fig:subfig-c}
  \end{subfigure}
  \vspace{-0.3cm}

  \caption{Plots of standard deviations for gradients of the supernet on CIFAR-10, CIFAR-100~\cite{krizhevsky2009cifar}, and ImageNet16-120~\cite{deng2009imagenet} in the NAS-Bench-201 space~\cite{dong2020bench}.}
  \label{fig:supp-noisy}
  \vspace{-0.5cm}
\end{figure*}

\begin{table}[t]
  \centering
  \small
  \caption{Quantitative comparisons of \text{CB} and \textit{$C^3$} on CIFAR-10, CIFAR-100~\cite{krizhevsky2009cifar}, and ImageNet16-120~\cite{deng2009imagenet} in the NAS-Bench-201 space~\cite{dong2020bench}.}
  \label{tab:cb_c3}

      \begin{tabular}{lcccccccc}
          \toprule
          \multirow{2}{*}{\vspace{-0.15cm} Method} & \multicolumn{2}{c}{CIFAR-10} & \multicolumn{2}{c}{CIFAR-100} & \multicolumn{2}{c}{ImageNet16-120} \\
          \cmidrule(lr){2-3} \cmidrule(lr){4-5} \cmidrule(lr){6-7}
          & CB & $C^3$ & CB & $C^3$ & CB & $C^3$ \\
          \midrule
          SPOS & 0.76 & -0.19 & 0.84 & -0.25 & 0.81  & -0.22 \\
          \rowcolor{gray!20}
          +CaLR & 0.63 & -0.09 & 0.7 & -0.11 & 0.68  & -0.1\\
          \midrule
          FairNAS & 0.72 & -0.13 & 0.69 & -0.16 & 0.73  & -0.15\\
          \rowcolor{gray!20}
          +CaLR & 0.62 & -0.06 & 0.6 & -0.1 &  0.65  & -0.08\\
          \midrule
          FSNAS & 0.79 & -0.21 & 0.82 & -0.24 & 0.8  & -0.22\\
          \rowcolor{gray!20}
          +CaLR & 0.63 & -0.1 & 0.66 & -0.11 &  0.65 & -0.05\\
          \hline
      \end{tabular}
  
\end{table}

\section{More Discussions}
\label{sec:app_discussion}
In this section, we provide additional discussions on the ablation studies and the design choices of our method.

\subsection{Ablations on CaLR and MS}

To further validate the effectiveness of our method in various NAS methods and datasets beyond Table~4 and Fig.~6 of the main paper, we show in Table~\ref{tab:cb_c3} and Fig.~\ref{fig:supp-noisy} additional ablation studies of CaLR and MS, conducted on various methods~(SPOS~\cite{guo2020single}, FairNAS~\cite{chu2021fairnas}, and FSNAS~\cite{zhao2021few}) and datasets~(CIFAR-10, CIFAR-100~\cite{krizhevsky2009cifar}, and ImageNet16-120~\cite{deng2009imagenet}) in the NAS-Bench-201 space~\cite{dong2020bench}. We can see that applying CaLR and MS alleviates the unfairness and noisy momentum problems, respectively, across different methods and datasets, demonstrating the generalizability of our method.

\begin{table}[t]
  \captionsetup{font={small}}
  \captionof{table}{Quantitative results of the ranking consistency on CIFAR-10~\cite{krizhevsky2009cifar} in the NAS-Bench-201 space~\cite{dong2020bench} with different choices of $g(x)$.}
\centering
\small
\resizebox{0.9\linewidth}{!}{
\begin{tabular}{ccccc}
\toprule
Baselines & CaLR & MS & $g(x)$ & Kendall's Tau \\
\midrule
\multirow{4}{*}{SPOS} & - & $\checkmark$ & - & $0.772\pm0.007$ \\
                      & $\checkmark$ &  $\checkmark$ & $e^x$ & $0.774\pm0.018$ \\
                      & $\checkmark$ &  $\checkmark$ & $x$ & $0.783\pm0.013$ \\
                      & $\checkmark$ &  $\checkmark$ & $\log(x)$ & $\textbf{0.814}\pm0.007$\\
\midrule                      
\multirow{4}{*}{FairNAS} & - & $\checkmark$ & - & $0.784\pm0.007$ \\
                      & $\checkmark$ &  $\checkmark$ & $e^x$ & $0.794\pm0.016$ \\
                      & $\checkmark$ &  $\checkmark$ & $x$ & $0.813\pm0.011$ \\
                      & $\checkmark$ &  $\checkmark$ & $\log(x)$ & $\textbf{0.828}\pm0.020$ \\

\midrule
\multirow{4}{*}{FSNAS} & - & $\checkmark$ & - & $0.750\pm0.024$ \\
                      & $\checkmark$ &  $\checkmark$ & $e^x$ & $0.755\pm0.015$ \\
                      & $\checkmark$ &  $\checkmark$ & $x$ & $0.763\pm0.013$ \\
                      & $\checkmark$ &  $\checkmark$ & $\log(x)$ & $\textbf{0.767}\pm0.010$ \\    
\bottomrule
\end{tabular}
}
\label{tab:gamma_ablation}
\end{table}

\begin{figure}[t]
\begin{center}
    \includegraphics[width=0.8\linewidth]{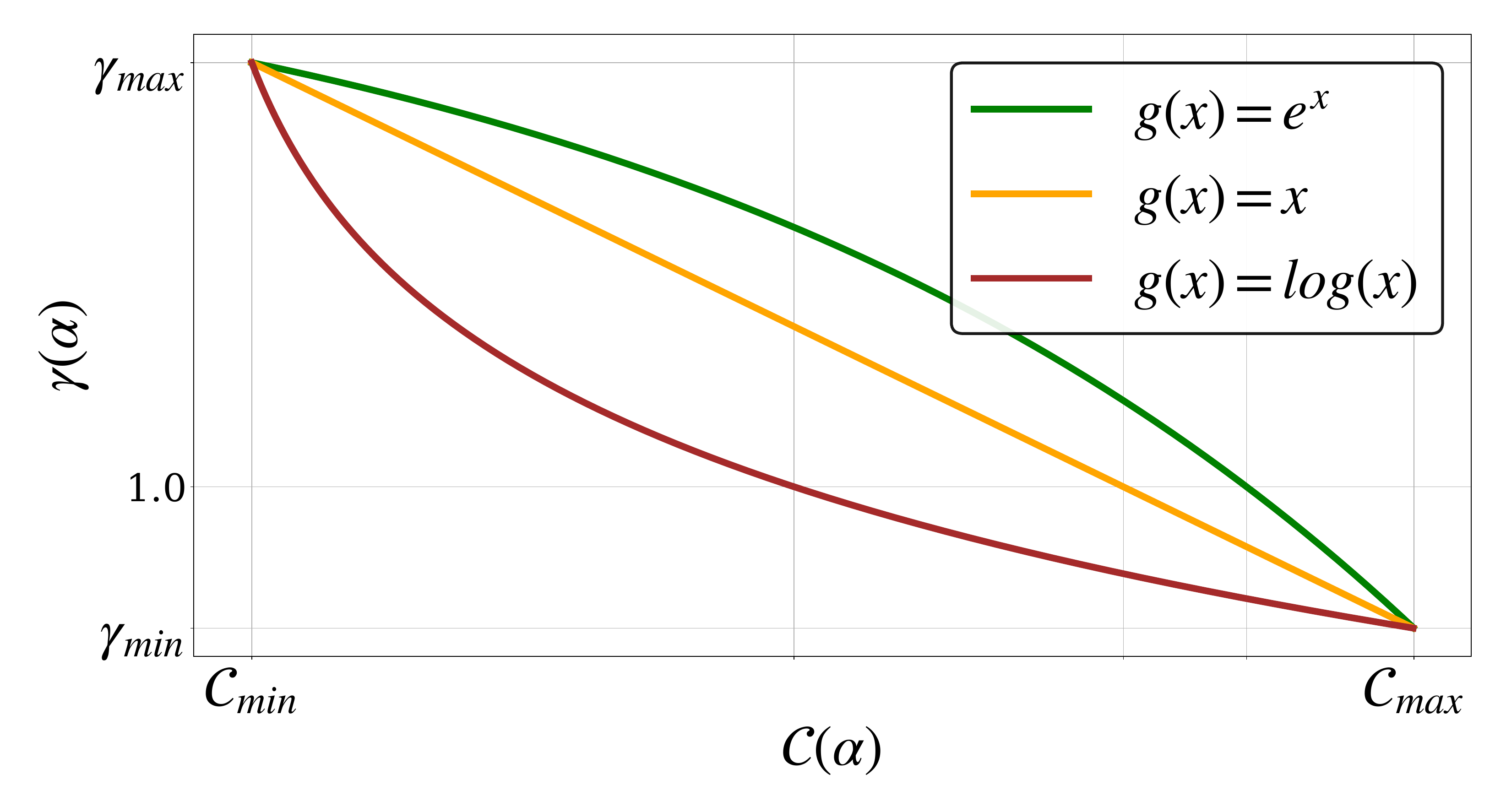}
\end{center}
\vspace{-0.4cm}
\captionsetup{font={small}}
   \caption{Visualization of the decay ratio $\gamma(\alpha)$ based on the complexity score $\mathcal{C}(\alpha)$ with different choices of $g(x)$.}
\label{fig:gamma_ablation}
\end{figure}

\subsection{Variants of the design of decay ratio in CaLR}
To compute the decay ratio $\gamma(\alpha)$ of a subnet $\alpha$, we first feed the complexity score $\mathcal{C}(\alpha)$ of the subnet $\alpha$ into a logarithmic function, followed by applying an affine transformation to $\log(\mathcal{C}(\alpha))$~(see Eq.~(4)). We consider the different design options for $\gamma(\alpha)$. To this end, we first reformulate Eq.~(4) as follows:
\begin{equation}
    \gamma(\alpha) = \omega g(\mathcal{C}(\alpha)) + \tau,
\end{equation}
where $\omega$ and $\tau$ are coefficients for the affine transformation:
\begin{flalign}
    \omega &= -\frac{\gamma_{\max} - \gamma_{\min}}{g(\mathcal{C}_{\max}) - g(\mathcal{C}_{\min})}, \\ 
    \tau &= \gamma_{\min} - \omega  g(\mathcal{C}_{\max}).  	
\end{flalign}
We show in Table~\ref{tab:gamma_ablation} the results of the ranking consistency on CIFAR-10~\cite{krizhevsky2009cifar} in the NAS-Bench-201 space~\cite{dong2020bench} with different functions for $g(x)$, including exponential $e^x$, linear $x$, and logarithmic $\log(x)$. From the results, we observe that $g(x) = \log(x)$ outperforms other options. Note that $g(x)$ determines the degree of the differentiation of the decay ratio $\gamma(\alpha)$ for the subnets with different complexities. The logarithmic function allows the subnet with a complexity score close to the medium~(around the center point between $\mathcal{C}_\text{min}$ and $\mathcal{C}_\text{max}$) to have a linear decaying LR, \ie, the decay ratio $\gamma(\alpha)$ of 1~(Fig.~\ref{fig:gamma_ablation}). This ensures LR of the high-complexity subnets to be slowly decayed~(\ie, $\gamma_\text{min}<\gamma(\alpha)<1$), while decaying it faster for low-complexity ones~(\ie, $1<\gamma(\alpha)<\gamma_\text{max}$). This suggests that designing $g(x)$ in a logarithmic function is effective in terms of differentiating the LR decay for various subnets. Other options, such as $x$, and $e^x$, fail to differentiate LR decay for varying complexities of subnets~(Fig.~\ref{fig:gamma_ablation}), since they do not satisfy the aforementioned criteria, resulting in an inferior performance. For these reasons, we adopt $g(x) = \log(x)$ for computing the decay ratio $\gamma(\alpha)$.

\begin{table}[t]
  \captionsetup{font={small}}
  \caption{Quantitative results of the ranking consistency on CIFAR-10~\cite{krizhevsky2009cifar} in the NAS-Bench-201 space~\cite{dong2020bench} with different choices of metrics for complexity. $\mathcal{C}(\alpha)$ denotes the complexity score of a subnet $\alpha$.}
  \centering
  \small
  \begin{tabular}{ccccc}
    \toprule
    \begin{tabular}[c]{@{}c@{}}Baselines\end{tabular} & CaLR & MS & $\mathcal{C}(\alpha)$ & \begin{tabular}[c]{@{}c@{}}Kendall's Tau\end{tabular} \\
    \midrule
    \multirow{3}{*}{SPOS} & - & - & - & $0.751\pm0.008$  \\
                          & $\checkmark$ & $\checkmark$ & Params & $\textbf{0.814}\pm0.007$ \\
                          & $\checkmark$ & $\checkmark$ & FLOPs & $0.783\pm0.015$  \\

    \midrule
    \multirow{3}{*}{FairNAS} & - & - & - & $0.766\pm0.015$  \\
                          & $\checkmark$ & $\checkmark$ & Params & $\textbf{0.828}\pm0.020$ \\
                          & $\checkmark$ & $\checkmark$ & FLOPs & $0.812\pm0.017$  \\

    \midrule
    \multirow{3}{*}{FSNAS} & - & - & - & $0.729\pm0.019$  \\
                          & $\checkmark$ & $\checkmark$ & Params & $\textbf{0.767}\pm0.010$ \\
                          & $\checkmark$ & $\checkmark$ & FLOPs & $0.756\pm0.010$  \\                      
  \bottomrule   
  \end{tabular}
  
  \label{tab:comp_met}
  \end{table}

\subsection{Metrics for complexity}
We show in Table~\ref{tab:comp_met} the results of the ranking consistency on CIFAR-10~\cite{krizhevsky2009cifar} in the NAS-Bench-201 space~\cite{dong2020bench} with different metrics for complexity. We consider the number of parameters~(Params), and FLOPs as metrics for complexity. We observe that using the number of parameters achieves slightly better performance than FLOPs. Moreover, since the number of parameters is a more straightforward to compute than FLOPs, which requires a forward pass, we adopt the number of parameters as a metric for complexity in our framework.

\begin{table}[t]
  \centering
  \footnotesize
  \captionsetup{font={small}} 
  \setlength{\tabcolsep}{3pt}
  \caption{Quantitative comparison of Kendall's Tau on CIFAR-10 in the NAS-Bench-201 space using different subnet sampling strategies and LR schedulers for supernet training. The subnet sampling strategies include uniform sampling and sampling proportional to the number of parameters (\#Params). The LR schedulers include the Cosine Annealing Scheduler (CS) and Complexity-aware LR scheduler (CaLR). The momentum separation technique~(MS) is used as a default for all methods. We report the average and standard deviations for 3 runs.}
  \begin{tabular}{lccc}
      \toprule
      Baselines & Subnet Sampling & \makecell{Scheduler \\ (Supernet Training)} & Kendall's Tau \\
      \midrule
      SPOS~\cite{guo2020single}      & Uniform      & CS    & 0.772 $\pm$ 0.007 \\
      SPOS~\cite{guo2020single}      & \#Params     & CS    & 0.784 $\pm$ 0.012 \\
      SPOS~\cite{guo2020single}     & Uniform      & CaLR  & \textbf{0.814} $\pm$ 0.007 \\
      \midrule
      FairNAS~\cite{chu2021fairnas}   & Uniform      & CS    & 0.784 $\pm$ 0.007 \\
      FairNAS~\cite{chu2021fairnas}    & \#Params     & CS    & 0.811 $\pm$ 0.013 \\
      FairNAS~\cite{chu2021fairnas}    & Uniform      & CaLR  & \textbf{0.828} $\pm$ 0.020 \\
      \midrule
      FSNAS~\cite{zhao2021few}    & Uniform      & CS    & 0.750 $\pm$ 0.024 \\
      FSNAS~\cite{zhao2021few}        & \#Params     & CS    & 0.761 $\pm$ 0.011 \\
      FSNAS~\cite{zhao2021few}       & Uniform      & CaLR  & \textbf{0.767} $\pm$ 0.010 \\
      \bottomrule
  \end{tabular}
  \label{tab:non_uniform}
\end{table}

\subsection{Using non-uniform sampling method in supernet training}
CaLR is designed to address the unfairness problem in supernet training by adjusting the LR based on the complexity of each subnet. One possible alternative to address the unfairness problem is to adjust the sampling probabilities of subnets based on their complexities. To compare these two approaches, we show in Table~\ref{tab:non_uniform} the results of using non-uniform sampling strategy, which samples the subnets with probabilities proportional to their complexities, instead of using CaLR in supernet training. We can see that adjusting the sampling probabilities improves the baselines, but shows inferior results compared to CaLR. Adjusting the sampling probabilities can cause a part of subnets to be sampled frequently while others are neglected, suggesting that large numbers of subnets would not be trained. In contrast, our CaLR maintains uniform sampling probabilities for all subnets but adjusts the LR based on the complexity of each subnet. This approach helps to balance the training across different subnets more effectively, ensuring that a wider range of subnets receive sufficient training.

\begin{table}[t]
  \centering
  \captionsetup{font={small}} 
  \setlength{\tabcolsep}{10pt}
  \small
  \caption{Quantitative results of top-1 accuracies of the searched subnets on CIFAR-10 in the NAS-Bench-201 space using different schedulers for the retraining stage. The schedulers include the Cosine Annealing Scheduler (CS) and Complexity-aware LR scheduler (CaLR). For supernet training, we use consistent settings, applying CaLR and MS to baselines. We report the average and standard deviations for 3 runs.}
  \begin{tabular}{lcc}
      \toprule
      Baselines & \makecell{Scheduler \\ (Retraining)} & Top-1 Acc. \\
      \midrule
      SPOS~\cite{guo2020single}   & CS   & \textbf{93.50} $\pm$ 0.33 \\
      SPOS~\cite{guo2020single}      & CaLR & 90.51 $\pm$ 0.48 \\
      \midrule
      FairNAS~\cite{chu2021fairnas}  & CS   & \textbf{93.52} $\pm$ 0.50 \\
      FairNAS~\cite{chu2021fairnas}   & CaLR & 90.75 $\pm$ 0.27 \\
      \midrule
      FSNAS~\cite{zhao2021few}    & CS   & \textbf{93.63} $\pm$ 0.21 \\
      FSNAS~\cite{zhao2021few}     & CaLR & 90.24$\pm$ 0.47 \\
      \bottomrule
  \end{tabular}
  \label{tab:calr_in_retrain}
\end{table}

\subsection{Using CaLR in the retraining stage}
Once we train a supernet with CaLR, and search the subnets from the trained supernet, we retrain the searched subnets with a standard scheduler~(\ie, cosine scheduler). To investigate the impact of using CaLR during retraining phase, we compare in Table~\ref{tab:calr_in_retrain} quantitative results between using CaLR and a standard scheduler in the retraining phase. We can see that using CaLR during a retraining stage does not improve performance compared to using a regular scheduler. The primary reason is that retraining aims to optimize a single subnet, and the cosine annealing scheduler is effective for this purpose, as it gradually lowers the LR, allowing the model to converge smoothly to a local minimum. In contrast, CaLR adjusts LRs based on the complexities of subnets, \eg, keeping high LR for high-complexity subnets, which is effective for supernet training to address the unfairness problem, but not optimal for retraining a single subnet.

To explain in more detail, our CaLR method is specifically designed to address the challenges of supernet training, where the main difficulty stems from the vast number of subnets within the search space (\eg, $\sim7^{21}$ for the MobileNet space). In this scenario, fully training each subnet to convergence within a limited number of iterations is impractical. Instead of extending the training iterations across all subnets, CaLR adjusts the LR dynamically based on the complexity of each subnet. This approach ensures that higher-complexity subnets, which have more parameters to optimize, receive relatively more training amounts compared to lower-complexity subnets. By doing so, CaLR balances the training process by giving more attention to subnets that need it due to their inherent complexity, alleviating the unfairness problem in supernet training.

In contrast, the retraining phase operates under different conditions where each subnet can be trained to full convergence. Applying CaLR during retraining may lead to suboptimal performance: maintaining a high LR for high-complexity subnets could result in overshooting, which hampers convergence, while a rapidly decaying LR for low-complexity subnets might require more iterations to achieve convergence. Hence, while CaLR is effective for supernet training, it may not be suitable for the retraining phase where a gradual LR decay, like the cosine annealing scheduler, is more appropriate.

\begin{figure}[t]
  \begin{center}
    \includegraphics[width=\linewidth]{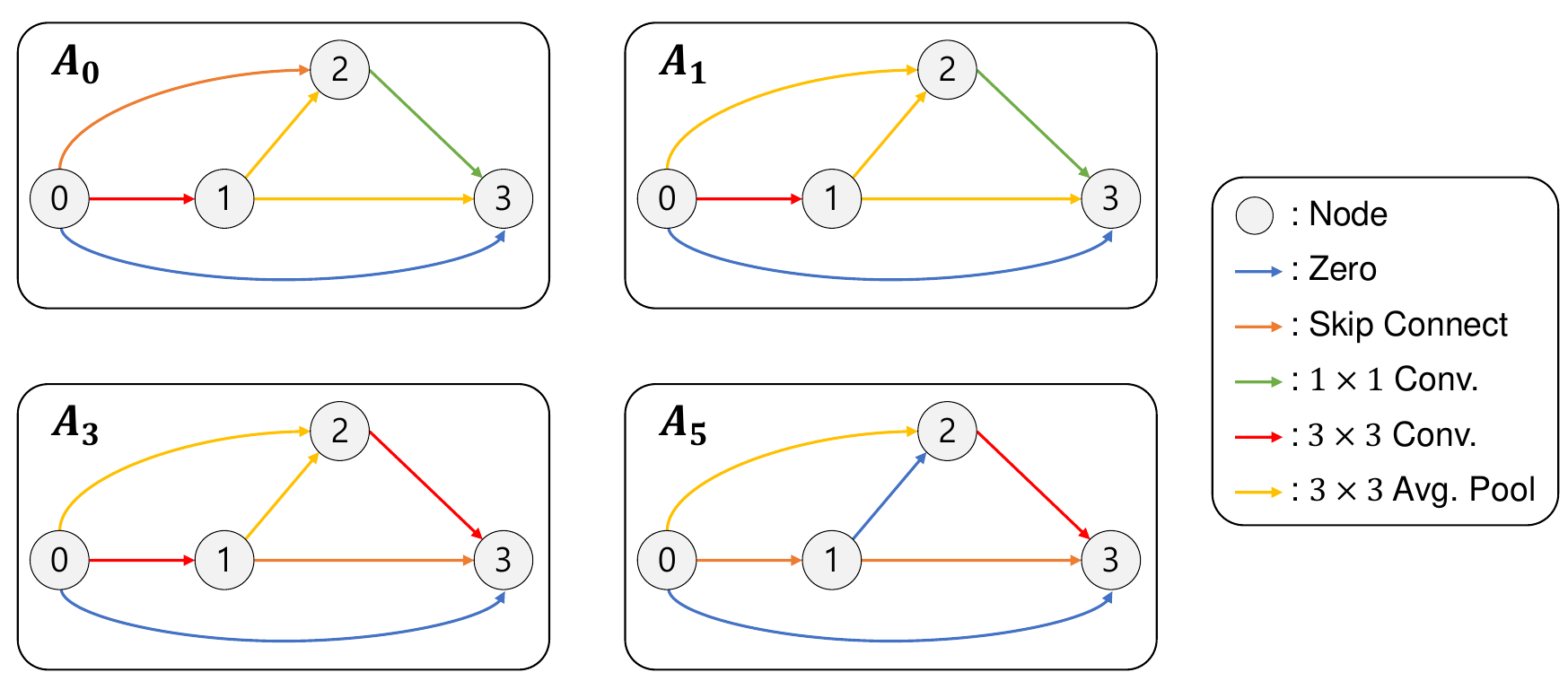}
\end{center}
\vspace{-0.4cm}
\captionsetup{font={small}}
   \caption{Visualization of the cell architectures of the generated subnets.}
\label{fig:generated_subnets}
\end{figure}

\begin{table}[t]

\captionsetup{font={small}}
  \captionof{table}{Quantitative results of mean gradient cosine similarity (MGCS) with respect to $A_0$ on CIFAR-10 in NAS-Bench-201.}
\centering
\small

  \begin{tabular}{cc}
    \toprule
    Subnet & MGSC \\
    \midrule
    $A_1$ & \textbf{0.6} \\
    $A_3$ & 0.27 \\
    $A_5$ & 0.09 \\
    \bottomrule
    \end{tabular}

\label{tab:MGSC}
    
\end{table}

 \subsection{Relationship between structural similarity and gradient consistency} 
We have shown that our MS, which clusters subnets based on their structural similarities, provides better results in terms of gradient consistencies, compared to the baseline NAS methods and a random clustering approach, in Fig.~2(c) and Fig.~6 of the main paper. This demonstrates that subnets with similar structures tend to make more consistent gradients.

 To further validate the statement, we perform a quantitative evaluation by computing cosine similarities between gradients, while varying network structures in terms of structural similarities in NAS-Bench-201. Specifically, we select the base subnet $A_0$. We then randomly change the operations of $n$ edges. By setting $n$ as 1, 3, and 5, the generated subnets are shown in Fig.~\ref{fig:generated_subnets}. Note that $A_1$ is the most similar network for $A_0$ in terms of the structural similarity, while $A_5$ is the most distinct one.

  The supernet, consisting of subnets $A_0$, $A_1$, $A_3$, and $A_5$, is trained using the accumulated gradients from all subnets at each iteration. During the last epoch of training, we compute the cosine similarities between gradients of the first layer of the subnets for every iteration. Specifically, we compute the similarities between the base network $A_0$ and $A_n$, $n=1, 3, 5$, \ie, $cos(g(A_0), g(A_n))$, where $cos()$ and $g(A_n)$ represent the cosine similarities and the gradients of the first layer of subnet $A_n$, respectively. We then report the mean gradient cosine similarities (MGCS), by averaging $cos(g(A_0), g(A_n))$ for all iterations. We show in Table~\ref{tab:MGSC} the results of MGSC. We can observe that subnets with higher structural similarities to $A_0$ show higher MGCS, suggesting that subnets with similar structures tend to generate similar gradients.

  \subsection{Impact of excluding momentum in supernet training}
  We have shown that sharing a single optimizer for all subnets makes momentum updates noisy. The natural question is that what the impact of excluding momentum in supernet training~(\ie, $\beta=0$ in Eq.~(8)) would be. To this end, we compare the performance of the baselines under three different conditions: without momentum ($\beta=0$), with momentum ($\beta=0.9$), and with MS in Table~\ref{tab:momentum}. We observe that the baseline without momentum achieves inferior performance compared to the baseline with momentum.  This indicates that despite the noisy nature of momentum updates in supernets, momentum still plays a crucial role in stabilizing the optimization process. Utilizing MS with momentum, however, can achieve better performance, since it can reduce the noise in momentum updates by clustering the subnets with similar structural characteristics. We also observe that the performance degradation of FairNAS~\cite{chu2021fairnas} without momentum is smaller than that of SPOS~\cite{guo2020single} and FSNAS~\cite{zhao2021few}. The reason for this is the design of FairNAS, which involves the use of multiple subnets in each iteration. The process of averaging the gradients from these subnets effectively mimics the smoothing effect of momentum, thereby mitigating the impact of its absence and resulting in a smaller performance drop for FairNAS without momentum.
  
  \begin{table}[t]
    \captionsetup{font={small}}
    \caption{Quantitative results of the ranking consistency on CIFAR-10~\cite{krizhevsky2009cifar} in the NAS-Bench-201 space~\cite{dong2020bench}. We compare our framework with different momentum settings.}
    \label{tab:momentum}
    \centering
    \small
    \begin{tabular}{cccc}
      \toprule
      \begin{tabular}[c]{@{}c@{}}Baselines\end{tabular} & MS & $\beta$ & \begin{tabular}[c]{@{}c@{}}Kendall's Tau\end{tabular} \\
      \midrule
      \multirow{3}{*}{SPOS} & - & 0 & $0.615\pm0.046$  \\
                            & - & 0.9 & $0.751\pm0.008$  \\
                            & $\checkmark$ & 0.9 & $\textbf{0.772}\pm0.007$  \\
  
      \midrule
      \multirow{3}{*}{FairNAS} & - & 0 & $0.743\pm0.022$  \\
                            & - & 0.9 & $0.766\pm0.015$  \\
                            & $\checkmark$ & 0.9 & $\textbf{0.784}\pm0.007$  \\
  
      \midrule
      \multirow{3}{*}{FSNAS} & - & 0 & $0.616\pm0.016$  \\
                            & - & 0.9 & $0.729\pm0.019$  \\
                            & $\checkmark$ & 0.9 & $\textbf{0.750}\pm0.024$  \\
    \bottomrule   
    \end{tabular}
    
  \end{table}

  \begin{table}[t]
    \captionsetup{font={small}}
    \captionof{table}{Quantitative results of the ranking consistency on CIFAR-10~\cite{krizhevsky2009cifar} in the NAS-Bench-201 space~\cite{dong2020bench} with different $\gamma'$.}
    \centering
    \small
    \begin{tabular}{ccccc}
      \toprule
      \begin{tabular}[c]{@{}c@{}}Baseline\end{tabular} & CaLR & MS & $\gamma'$ & \begin{tabular}[c]{@{}c@{}}Kendall's Tau\end{tabular} \\
      \midrule
      \multirow{4}{*}{SPOS} & - & - & - & $0.751\pm0.008$  \\
                            & $\checkmark$ & $\checkmark$ & 2 & $0.787\pm0.001$ \\
                            & $\checkmark$ & $\checkmark$ & 3 & $0.797\pm0.010$  \\
                            & $\checkmark$ & $\checkmark$ & 4 & $\textbf{0.814}\pm0.007$  \\
    \bottomrule   
    \end{tabular}
    
    \label{tab:hyper_nas}
  \end{table}
  \begin{table}[t]
    \captionsetup{font={small}}
    \captionof{table}{Quantitative results of the search performance on ImageNet~\cite{deng2009imagenet} in the MobileNet space~\cite{cai2018proxylessnas} with different $\gamma'$.}
    \centering
    \small
    \begin{tabular}{ccccc}
      \toprule
      \begin{tabular}[c]{@{}c@{}}Baseline\end{tabular} & CaLR & MS & $\gamma'$ & \begin{tabular}[c]{@{}c@{}}Top-1 (\%)\end{tabular} \\
      \midrule
      \multirow{4}{*}{SPOS-S} & - & - & - & $74.3$  \\
                            & $\checkmark$ & $\checkmark$ & 2 & $74.4$ \\
                            & $\checkmark$ & $\checkmark$ & 3 & $\textbf{74.6}$  \\
                            & $\checkmark$ & $\checkmark$ & 4 & $74.5$ \\
    \bottomrule   
    \end{tabular}
    
    \label{tab:hyper_mobile}
\end{table}
  
  \subsection{Hyperparameter search}
  We show in Tables~\ref{tab:hyper_nas} and \ref{tab:hyper_mobile} the results of the hyperparameter search on NAS-Bench-201~\cite{dong2020bench} and MobileNet~\cite{cai2018proxylessnas} spaces\footnote{We retrain the searched subnet in MobileNet space for 240 epochs when searching hyperparameters, for efficiency.}, respectively. Our framework has two hyperparameters, including the maximum and minimum decay ratio of CaLR, $\gamma_\text{max}$ and $\gamma_\text{min}$, respectively, and we set $\gamma_\text{max} = 1/\gamma_\text{min}=\gamma'$ for simplicity. We perform the grid search to determine the maximum and minimum decay ratio of CaLR; $\gamma' \in \{2,3,4\}$. From the results, we observe that our framework achieves the best performance when $\gamma' = 4$ and $3$ for NAS-Bench-201 and MobileNet spaces, respectively. This variation in optimal hyperparameter settings can be attributed to the differences in the distribution of the number of parameters among subnets across these search spaces. Furthermore, our framework consistently outperforms the baselines across various hyperparameter configurations, demonstrating its robustness and adaptability to different hyperparameter choices.

\subsection{Layer selection in MS}
We show in Table~\ref{tab:layer} an ablation study on layer selection for MS in the MobileNet space. We compare two cases for selecting a layer in MS~(the first layer and the last layer). We observe that choosing the first layer achieves superior performance compared to selecting the last layer. We attribute this to the distinct characteristics of the layers in the subnets. As suggested in~\cite{peng2021pi}, the feature distributions of each layer highly affect the gradient flow in the subnet. The features in the deeper layers are likely to be more diverse across the subnets, due to the cumulative effect of operation choices made in all preceding layers. This diversity in feature distributions leads to more inconsistent gradients. Therefore, selecting the deep layer for MS may not significantly contribute to the stabilization of momentum updates. In contrast, sharing the same operation in the shallower layer results in more consistent feature distributions across the subnets, due to the property of the shallow layers that learn more general features such as edges and textures. Selecting the first layer for MS thus results in more consistent gradients within the cluster than selecting the deeper layers, which can stabilize the momentum updates.

\begin{table}[t]
  \captionsetup{font={small}}
  \caption{Quantitative results of the search performance on ImageNet~\cite{deng2009imagenet} in the MobileNet space~\cite{cai2018proxylessnas}. We compare the layer selection strategy in MS. We report the top-1 validation accuracy of the searched subnets.}
  \setlength{\tabcolsep}{0.3em}
  \centering
  \small
  \begin{tabular}{ccccc}
    \toprule
    \begin{tabular}[c]{@{}c@{}}Baseline\end{tabular} & CaLR & MS & Layer selection & \begin{tabular}[c]{@{}c@{}}Top-1 (\%)\end{tabular} \\
    \midrule
    \multirow{3}{*}{SPOS-S} & - & - & - & 74.6  \\
                          & $\checkmark$ & $\checkmark$ & First layer & \textbf{74.8}\\
                          & $\checkmark$ & $\checkmark$ & Last layer & 74.7  \\
  \bottomrule   
  \end{tabular}
  
  \label{tab:layer}
\end{table}

  \section{Extensions to various \textit{N}-shot NAS methods}
  \label{sec:app_algorithm}
  We provide overall processes of applying our framework on various NAS methods, including SPOS~\cite{guo2020single}, FairNAS~\cite{chu2021fairnas}, and FSNAS~\cite{zhao2021few} in Algorithms~\ref{alg:spos},~\ref{alg:fair}, and~\ref{alg:few}, respectively. In the following, we detail the extensions to various NAS methods other than SPOS.
  
  \paragraph{FairNAS.} For FairNAS, we sample multiple subnets at each iteration, and compute the gradients of each subnet. We then update the parameters through a weighted sum of the gradients, where the weights assigned to each gradient are determined based on the corresponding subnet using Eq.~(4) of the main paper. This approach ensures that the influence of each subnet's gradient on the overall update is proportional to their complexities. Additionally, to facilitate the update of momentum with the accumulated gradients, we sample the subnets from the same cluster at each iteration.  
  
  \paragraph{FSNAS.} FSNAS utilizes multiple sub-supernets partitioned from a supernet. We apply SPOS~\cite{guo2020single} with our dynamic supernet training framework to optimize each sub-supernet. Unlike SPOS and FairNAS, FSNAS retrains the retrieved subnet of each sub-supernet and selects the best subnet among them.

  \section{Subnets} 
  \label{sec:app_searched_subnets}
  For reproducibility, we provide in Figs.~\ref{fig:spos},~\ref{fig:FairNAS}, and~\ref{fig:FSNAS} the searched subnets on ImageNet~\cite{deng2009imagenet} in the MobileNet space~\cite{cai2018proxylessnas} using SPOS~\cite{guo2020single} + Ours, FairNAS~\cite{chu2021fairnas} + Ours, and FSNAS~\cite{zhao2021few} + Ours as search algorithms, respectively. Note that MB$x$ $k\times k$ represents a MobileNet block with an expansion ratio of $x$ and a kernel size of $k\times k$. The numbers over the blocks denote the number of output channels.

\renewcommand{\algorithmiccomment}[1]{#1}
\vspace{3cm}
      \begin{algorithm*}[h]
        \caption{Dynamic supernet training on SPOS~\cite{guo2020single}}
        \label{alg:spos}
        \begin{algorithmic}[1]
            \STATE \textbf{Input}: Supernet $\mathcal{N}$, weights of supernet $\mathcal{W}$, momentum of supernet $\mu$, training set $\mathcal{D}_{train}$, total number of iterations $T$.
            \STATE Initialize each cluster's momentum: $\mu^0 = 0$.
            \FOR{$t=1$ to $T$}
                \STATE Sample a mini-batch from training set~$\mathcal{D}_{train}$.
                \STATE Randomly sample a subnet $\alpha$ from a supernet~$\mathcal{N}$.
                \STATE Compute LR $\eta^{t}$ of current iteration~$t$, considering a complexity score of the subnet~$\mathcal{C}(\alpha)$ using Eq.~(3).
                \STATE Obtain momentum $\mu_i$ from cluster $i$ where subnet $\alpha$ is located using Eq.~(7).
                \STATE Compute gradients $g^t$ of the subnet $\alpha$ \wrt train loss.
                \STATE Update momentum:
               $\mu_i^t = \beta \cdot \mu_i^{t-1} + g^t$, 
                  where $\beta$ is a coefficient of moving average.
                \STATE Update weights:
                  $\mathcal{W}^t(\alpha) = \mathcal{W}^{t-1}(\alpha) - \eta^{t} \cdot \mu_i^t$.
            \ENDFOR
            
        \end{algorithmic}
      \end{algorithm*}
  
      \begin{algorithm*}[h]
          \caption{Dynamic supernet training on FairNAS~\cite{chu2021fairnas}}
          \label{alg:fair}
          \begin{algorithmic}[1]
              \STATE \textbf{Input}: Supernet $\mathcal{N}$, weights of supernet $\mathcal{W}$, momentum of supernet $\mu$, set of clusters $S$, supernet depth $L$, number of candidate operations $n$, training set $\mathcal{D}_{train}$, total number of iterations $T$.
              \STATE Initialize each cluster's momentum: $\mu^0 = 0$.
              \FOR{$t=1$ to $T$}
                  \STATE Sample a mini-batch from training set~$\mathcal{D}_{train}$.
                  \STATE Randomly select a cluster $S_i$ from the set of clusters $S$.
                  \FOR{$l=1$ to $L$}
                      \STATE $c_l =$ a uniform permutation of index for $n$ candidate operations of layer $l$. 
                  \ENDFOR
                  \STATE Initialize gradients: $g^t = 0$.
                  \FOR{$k=1$ to $n$}
                      \STATE $c_{e_k} = i$, where $e$ is a selected layer for MS. \textcolor{gray}{// To ensure that the sampled subnets are within the cluster.}
                      \STATE Sample a subnet $\alpha_k$ = ($c_{1_k}, c_{2_k}, \cdots, c_{L_k}$).
                      \STATE Compute $\eta^{t}_k$ of current iteration~$t$, considering a complexity score of the subnet~$\mathcal{C}(\alpha_k)$ using Eq.~(3).
                      \STATE Compute gradients $g^t_k$ of the subnet $\alpha_k$ w.r.t train loss.
                      \STATE Accumulate gradients: $g^t \leftarrow g^t + g^t_k \cdot \eta^{t}_k$.
                  \ENDFOR
                  \STATE Update momentum:
                 $\mu_i^t = \beta \cdot \mu_i^{t-1} + g^t$, 
                    where $\beta$ is a coefficient of moving average.
                  \STATE Update weights:
                    $\mathcal{W}^t(\alpha) = \mathcal{W}^{t-1}(\alpha) - \mu_i^t$.
              \ENDFOR
              
          \end{algorithmic}
        \end{algorithm*}
      
      \begin{algorithm*}[h]
      \caption{Dynamic supernet training on FSNAS~\cite{zhao2021few}}
      \label{alg:few}
      \begin{algorithmic}[1]
          \STATE \textbf{Input}: Supernet $\mathcal{N}$, weights of supernet $\mathcal{W}$, momentum of supernet $\mu$, training set $\mathcal{D}_{train}$, total number of iterations $T$.
          \STATE Randomly split supernet $\mathcal{N}$ into $K$ sub-supernets: $\{\mathcal{N}_1, \mathcal{N}_2, \cdots, \mathcal{N}_K\}$.
          \FOR{$k=1$ to $K$}
          \STATE Initialize each cluster's momentum within a sub-supernet $\mathcal{N}_k$: $\mu^0_k = 0$.
          \FOR{$t=1$ to $T$}
              \STATE Sample a mini-batch from training set~$\mathcal{D}_{train}$.
              \STATE Randomly sample a subnet $\alpha$ from a supernet~$\mathcal{N}_k$.
              \STATE Compute LR $\eta^{t}$ of current iteration~$t$, considering a complexity score of the subnet~$\mathcal{C}(\alpha)$ using Eq.~(3).
              \STATE Obtain momentum $\mu_{k_i}$ from cluster $k_i$ where $\alpha$ is located using Eq.~(7).
              \STATE Compute gradients $g^t$ of the subnet $\alpha$ w.r.t train loss.
              \STATE Update momentum:
              $\mu_{k_i}^t = \beta \cdot \mu_{k_i}^{t-1} + g^t$, 
                  where $\beta$ is a coefficient of moving average.
              \STATE Update weights:
                  $\mathcal{W}_k^t(\alpha) = \mathcal{W}_k^{t-1}(\alpha) - \eta^{t} \cdot \mu_{k_i}^t$.
          \ENDFOR
          \ENDFOR

      \end{algorithmic}
      \end{algorithm*}

      

      \begin{figure*}[t]
        \captionsetup{font={small}}
        \begin{center}
          \begin{subfigure}{\linewidth}
               \centering
               \includegraphics[width=0.95\linewidth]{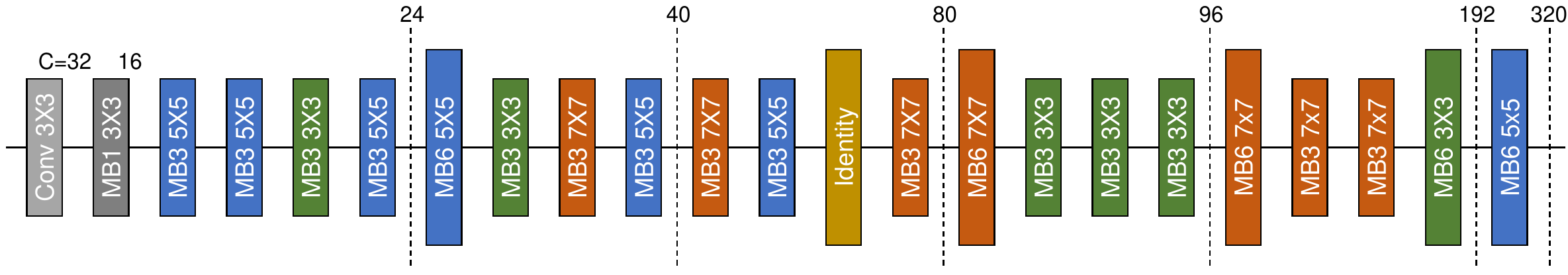}
               \caption{SPOS-S + Ours.}
          \end{subfigure}
          \begin{subfigure}{\linewidth}
            \centering
            \includegraphics[width=0.95\linewidth]{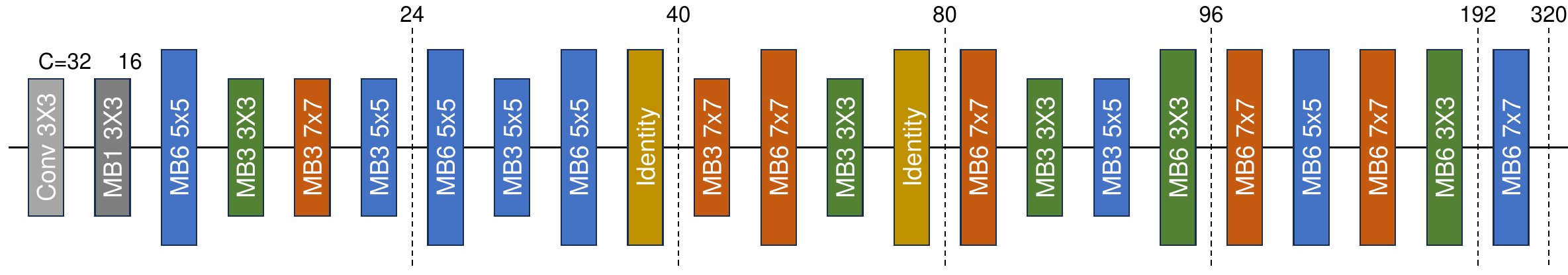}
            \caption{SPOS-M + Ours.}

        \end{subfigure}
        \begin{subfigure}{\linewidth}
            \centering
            \includegraphics[width=0.95\linewidth]{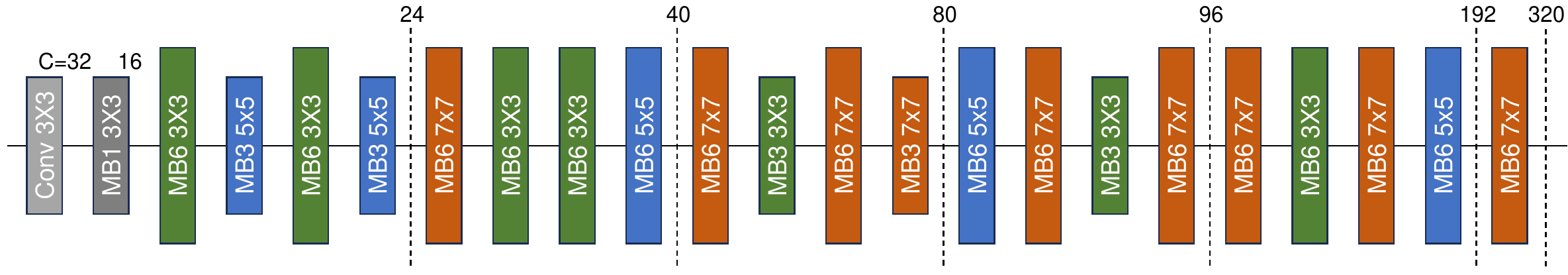}
            \caption{SPOS-L + Ours.}
       \end{subfigure}
      \end{center}
      \vspace{-0.4cm}
         \caption{Visualizations of our searched subnets on ImageNet~\cite{deng2009imagenet} in the MobileNet space~\cite{cai2018proxylessnas} using SPOS~\cite{guo2020single} + Ours.}
         \label{fig:spos}
       \end{figure*}
       \begin{figure*}[t]
        \captionsetup{font={small}}
        \begin{center}
          \begin{subfigure}{\linewidth}
               \centering
               \includegraphics[width=0.95\linewidth]{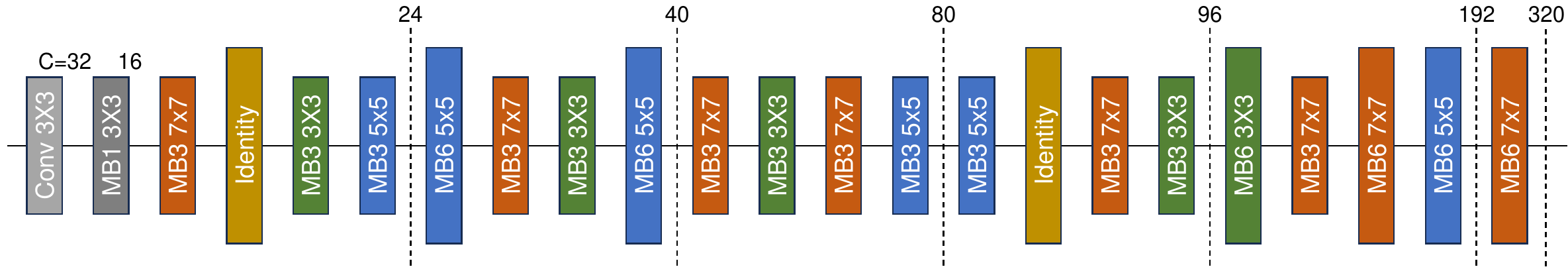}
               \vspace{-0.2cm}
               \caption{FairNAS-S + Ours.}
          \end{subfigure}
          \begin{subfigure}{\linewidth}
            \centering
            \includegraphics[width=0.95\linewidth]{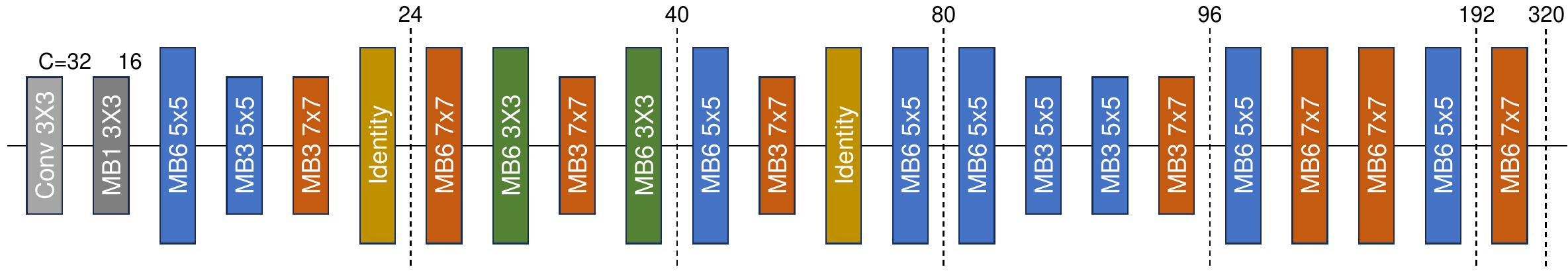}
            \vspace{-0.2cm}
            \caption{FairNAS-M + Ours.}

        \end{subfigure}
        \begin{subfigure}{\linewidth}
            \centering
            
            \includegraphics[width=0.95\linewidth]{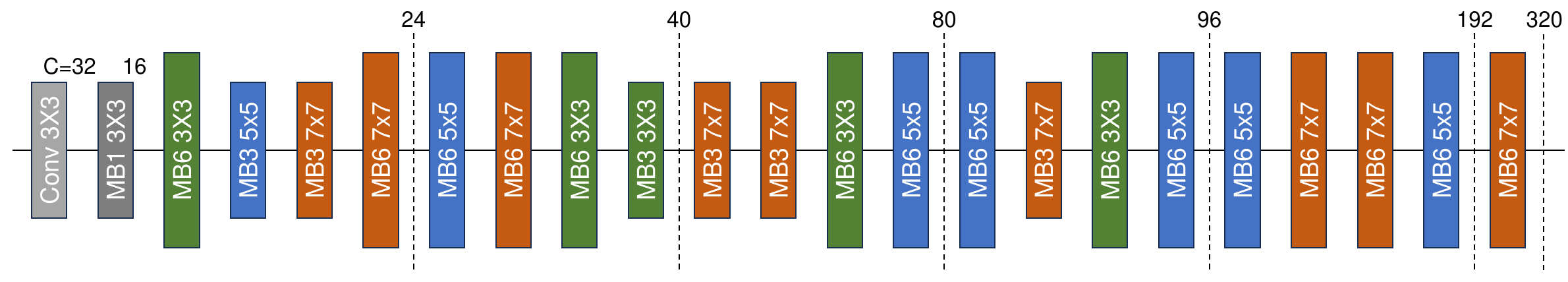}
            \vspace{-0.2cm}
            \caption{FairNAS-L + Ours.}
       \end{subfigure}
      \end{center}
      \vspace{-0.4cm}
         \caption{Visualizations of our searched subnets on ImageNet~\cite{deng2009imagenet} in the MobileNet space~\cite{cai2018proxylessnas} using FairNAS~\cite{chu2021fairnas} + Ours.}
         \label{fig:FairNAS}
       \end{figure*}

       \begin{figure*}[t]
        \captionsetup{font={small}}
        \begin{center}
          \begin{subfigure}{\linewidth}
               \centering
               \includegraphics[width=0.98\linewidth]{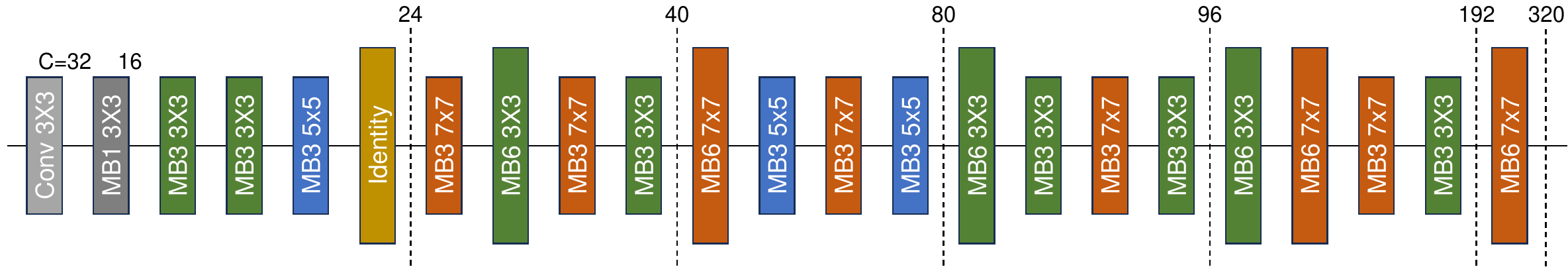}
               \caption{FSNAS-S + Ours.}
          \end{subfigure}
          \begin{subfigure}{\linewidth}
            \centering
            \includegraphics[width=0.98\linewidth]{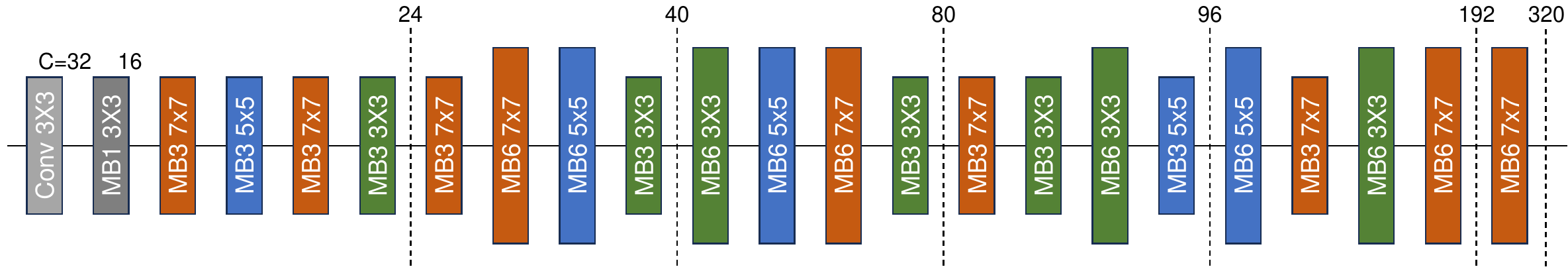}
            \caption{FSNAS-M + Ours.}

        \end{subfigure}
        \begin{subfigure}{\linewidth}
            \centering
            \includegraphics[width=0.98\linewidth]{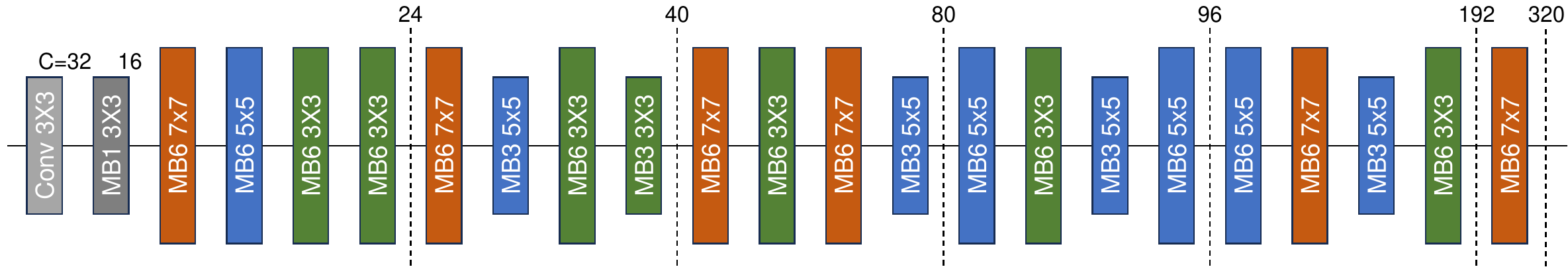}
            \caption{FSNAS-L + Ours.}
       \end{subfigure}
      \end{center}
      \vspace{-0.4cm}
         \caption{Visualizations of our searched subnets on ImageNet~\cite{deng2009imagenet} in the MobileNet space~\cite{cai2018proxylessnas} using FSNAS~\cite{zhao2021few} + Ours.}
         \label{fig:FSNAS}
       \end{figure*}

\clearpage
\clearpage
{
    \small
    \bibliographystyle{ieeenat_fullname}
    \bibliography{suppl}
}



\end{document}